\documentclass[preprint,12pt]{elsarticle}
\usepackage[a4paper, margin=1.5in]{geometry}

\usepackage{amssymb}
\usepackage{amsmath}

\usepackage{algpseudocode}
\usepackage{algorithm}

\usepackage{epstopdf}
\usepackage{subfigure}
\usepackage{amsthm}
\theoremstyle{plain}

\theoremstyle{definition}
\newtheorem{definition}{Definition}

\theoremstyle{remark}

\usepackage{comment}
\usepackage{hyperref}
\hypersetup{
    colorlinks=true,       
    linkcolor=red,          
    citecolor=blue,        
    filecolor=magenta,      
    urlcolor=cyan           
}

\usepackage[dvipsnames]{xcolor} 
\definecolor{colour1}{RGB}{166,206,227}
\definecolor{colour2}{RGB}{31,120,180}
\definecolor{colour3}{RGB}{178,55,250} 
\definecolor{colour4}{RGB}{51,160,44}

\usepackage{subcaption}
\usepackage{amsmath, amssymb}
\usepackage{amsthm}
\usepackage{bm}

\definecolor{purple}{RGB}{178,55,250} %

\newcounter{noteMCctr} 

\newcounter{noteASctr} 

\newcounter{noteBGctr} 

\usepackage{graphicx} 
\usepackage{adjustbox}

\makeatletter
\def\ps@pprintTitle{%
 \let\@oddhead\@empty
 \let\@evenhead\@empty
 \let\@oddfoot\@empty
 \let\@evenfoot\@empty
}
\begin{document}

\begin{frontmatter}



\title{Generalized Factor Neural Network Model for High-dimensional Regression}


\author[oxstat,omi]{Zichuan Guo} 
\author[oxstat,omi,ucla]{Mihai Cucuringu} 
\author[qmmath,nlmathstat]{Alexander Y. Shestopaloff} 

\affiliation[oxstat]{organization={Department of Statistics, University of Oxford},
            addressline={24-29 St Giles'}, 
            city={Oxford},
            postcode={OX1 3LB}, 
            country={U.K.}}

\affiliation[omi]{organization={Oxford-Man Institute, University of Oxford},
            addressline={Eagle House, Walton Well Road}, 
            city={Oxford},
            postcode={OX2 6ED}, 
            country={U.K.}}

\affiliation[ucla]{organization={Department of Mathematics, University of California Los Angeles},
            addressline={UCLA Mathematical Sciences Building, 520 Portola Plaza}, 
            city={Los Angeles},
            postcode={90095}, 
            country={US}}
            
\affiliation[qmmath]{organization={School of Mathematical Sciences, Queen Mary University of London},
            addressline={Mile End Road}, 
            city={London},
            postcode={E1 4NS},
            country={U.K.}}

\affiliation[nlmathstat]{organization={Department of Mathematics and Statistics, Memorial University of Newfoundland},
            addressline={230 Elizabeth Avenue}, 
            city={St. John's},
            postcode={A1C 5S7},
            state={NL},
            country={Canada}}

\begin{abstract}
We tackle the challenges of modeling high-dimensional data sets, particularly those with latent low-dimensional structures hidden within complex, non-linear, and noisy relationships. Our approach enables a seamless integration of concepts from non-parametric regression, factor models, and neural networks for high-dimensional regression. Our approach introduces PCA and Soft PCA layers, which can be embedded at any stage of a neural network architecture, allowing the model to alternate between factor modeling and non-linear transformations. This flexibility makes our method especially effective for processing hierarchical compositional data. We explore ours and other techniques for imposing low-rank structures on neural networks and examine how architectural design impacts model performance. The effectiveness of our method is demonstrated through simulation studies, as well as applications to forecasting future price movements of equity ETF indices and nowcasting with macroeconomic data.  
\end{abstract}


\begin{keyword}


High-dimensional Regression\sep Generalized Additive Model\sep Factor Model\sep Neural Network\sep Hierarchical Composition Model\sep Generalized PCA
\end{keyword}

\end{frontmatter}


\section{Introduction}
\label{sec:section1}

In the age of big data, an increasing amount of high-dimensional data is becoming available. Additionally, there is a growing need to make predictions based on a large number of variables \citep{Fan_2014, wainwright2019high}. As an illustration, consider the ImageNet data set \citep{deng2009large}, which consists of approximately $1.5\times10^7$ labeled images with high resolution and a size of around $2 \times 10^5$. Associated with high-dimensional features is the dependence among variables, and many laws of nature and human societies admit certain sparse compositional structures \citep{dahmen2022compositional}, for example, natural language \citep{farkas1986varieties} and gene sequencing data \citep{quinn2018understanding}. In the financial domain, researchers frequently face the challenge of building predictive models from limited data or generating a multitude of linear and non-linear features from low-dimensional data for a regression task \citep{https://doi.org/10.1111/jofi.13298}.

These observations underscore the necessity of tailoring statistical and machine learning models to capture low-dimensional structures within a high-dimensional regime, specifically in the regime $p \gg \log n$. Classical statistical tools, such as additive models \citep{friedman1981projection} or linear factor models used in factor analysis \citep{fruchter1954introduction}, explicitly delineate the structure of the data-generating process. However, relationships between factors representing the sparse structure and the dependent and independent variables, often extend beyond simple linearity. 

Considering the task of forecasting the returns of the S\&P500 index based on prices of its stock constituents, one may expect there exist factors that non-linearly affect the observed variables. \Citet{MCMILLAN2001353} posits a non-linear relationship between interest rates and stock returns. \Citet{reddy2019impact} demonstrate the asymmetric market response to credit rating changes, with more significant reactions to downgrades than upgrades. Similarly, \Citet{bernard1989post} reports a stronger market reaction to negative earnings surprises than positive ones. Additionally, we can also expect factors that have diminishing marginal impact --- fundamental factors like capital expenditures (CapEx) exhibit non-linear relationships with stock returns; initial CapEx investments can significantly enhance performance and competitive advantage, but the benefits diminish with subsequent investments due to saturation effects or strategic misalignments \citep{michelon2020capital}.

These phenomena suggest that although the data is governed by a few latent variables, the association between observed variables and latent factors can be non-linear. The versatility and superior performance of neural networks in handling non-linear patterns and complexities are well-documented \Citep{4341155, doi:10.1080/00031305.1996.10473554, somers2009using}. However, neural networks are data-intensive and may perform unstably in high-dimensional settings \Citep{saarinen1993ill}. This makes it imperative to seamlessly incorporate simple structures into neural network models, enhancing their efficiency while retaining their adaptability to non-linear dynamics. This can be achieved either by imposing low-dimensional structures on the data or on the regression function \citep{lu2023co, zhang2024graph}. In light of this, we will first provide an overview of the key concepts and methods that will be used to better address the high-dimensional regression problem.  

\subsection{Nonparametric Regression}
\label{subsec:nonpara}
Considering a covariate vector $x \in \mathbb{R}^p$ and a response variable $y$, originating from an unspecified distribution $\mu$, our objective revolves around the estimation of the regression function $m^*(\boldsymbol{x}) = \mathbb{E}[y \mid \boldsymbol{x}]$. This function is aimed at minimizing the population $L_2$ risk, defined as 

\begin{equation}
\mathrm{R}(m)=\int(y-m(x))^2 \mu(d x, d y)
\end{equation}

To address the challenges introduced by the \textit{curse of dimensionality}, adopting certain low-dimensional structures for the regression function $m^*(x)$ is a typical regularization strategy. Extensive research has been dedicated to characterizing inherent low-dimensional structures and devising efficient statistical approaches, including interaction models \citep{stone1994use}, and single-index models \citep{hardle1989investigating}. Implementing these low-dimensional frameworks proves beneficial in improving the rate of convergence. For instance, \citet{stone1985additive} applies an additive structure $m^*(x) = \sum_{j=1}^{p} m^*_j(x_j)$, demonstrating that a superior convergence rate of $n^{-2\beta/(2\beta+1)}$ is achievable with univariate functions that are $(\beta;C)$-smooth. In the nonlinear high-dimensional setting, \citet{scheidegger2023spectral} assumed an additive and sparse structure to develop a simple yet practical model.

\subsection{Factor Models}
Factor models are widely used in the analysis of high-dimensional data across various fields such as finance, economics, genomics, and social sciences. Factor analysis was developed by the British psychologist Charles Spearman in the early 20th century to analyze intelligence structures. Spearman's introduction of the general intelligence factor ("g") marked the beginning of factor model applications, showcasing their potential to reveal latent variables influencing observed outcomes \citep{YANAI2006257}. The statistical foundations of factor models were further solidified through efforts to mathematically construct factors for analysis, as seen in the work of \citet{krijnen2002construction}.

The primary motivation for using factor models is rooted in the idea that the observed variables in a data set may share common sources of variation that are not directly measured --- latent factors. Instead of treating each variable independently, factor models aim to capture the joint variability among variables by positing the existence of these latent factors. A linear factor model with covariate \( x \) admits
\begin{equation}
    x = Bz + u, 
    \label{eq:factor}
\end{equation}
where the latent factor \( z \in \mathbb{R}^r \) and the idiosyncratic component \( u \in \mathbb{R}^p \) is unobserved, the factor loading matrix \( B \in \mathbb{R}^{p \times r} \) is fixed but unknown. 


The literature on factor models is extensive, with numerous applications across finance. For example, the Fama-French Three-Factor Model extends the Capital Asset Pricing Model (CAPM) by incorporating size and value factors, providing a more comprehensive explanation of stock returns \citep{fama1992cross}. Similarly, the Carhart Four-Factor Model, which adds momentum as an additional factor, is widely used to explain mutual fund performance \citep{carhart1997persistence}. The Arbitrage Pricing Theory (APT), introduced by \citet{chen1986economic}, captures the influence of multiple macroeconomic factors on asset prices.

Recent developments include the asset pricing model proposed by \citet{kelly2019characteristics}, which assumes individual returns follow a 
$K-factor$ structure. Additionally, modern factor models play a critical role in estimating the covariance matrix of asset returns, a key component in portfolio construction and risk management \citep{zhang2023dynamic}. By reducing the dimensionality of the covariance matrix, factor models explain the co-movement of asset returns through a smaller set of common factors, improving estimation efficiency \citep{ledoit2003improved}.

\subsection{Neural Networks}

A neural network is a composition of multiple simple mathematical functions that implement more complex functions. \citet{cybenko1989approximation, hornik1991approximation,  pinkus1999approximation} showed that multi-layer perceptrons (i.e. feed-forward neural networks with at least one hidden layer) can approximate any continuous function, which is referred to as the universal approximation theorem. 

Recent progress in deep learning has led to significant breakthroughs in handling complex challenges, such as object detection from images and speech recognition \citep{lecun2015deep}. These tasks often involve analyzing data with a very high number of features relative to the number of samples available. This scenario underscores the capability of neural networks to manage high-dimensional data effectively. Notably, research by \citet{kohler2021rate} and \citet{schmidt2020nonparametric} has demonstrated that a deep ReLU neural network can adapt to the inherently low-dimensional structure of regression functions. This adaptability enables them to overcome the curse of dimensionality.

\subsection{The Problem Under Study}
Our research draws inspiration from the work of \citet{fan2023factor}, wherein they innovatively integrate factor models with neural network models through the use of diversified projections for estimating latent factor spaces. They leveraged deep ReLU networks for nonparametric factor regression, formulating a Factor Augmented Regression Using Neural Network (FAR-NN) estimator and a Factor Augmented Sparse Throughput Neural Network (FAST-NN) estimator. Their findings demonstrate that FAR-NN and FAST-NN estimators adeptly adjust to the underlying low-dimensional structures by employing hierarchical composition models, achieving nonasymptotic minimax rates.

In this methodology, the factor structure, commonly identified through methods such as PCA, is established in the initial step and subsequently integrated into the neural networks. This one-time factor identification process is particularly suited for observations with a low-dimensional data structure that can be modeled by a linear factor model, as shown in equation \eqref{eq:factor}.

Building on their contributions, our study seeks to broaden the applicability of this framework to high-dimensional datasets that exhibit more complex structures. Through this extension, we aim to uncover new insights and enhance the modeling precision for datasets characterized by intricate relationships and patterns. We posit that within these complex datasets, latent low-dimensional structures exist; however, it is conceivable that these low-dimensional structures might only manifest themselves following some non-linear transformations. We thus specify the observation follows a non-linear factor model with covariate \( x \) admits
\begin{equation}
    x = Bg(z) + u, 
    \label{eq:non-linear_factor}
\end{equation}
where \( g \) is a smooth, approximately monotonic, and non-degenerate non-linear function applied elementwise to \( \mathbf{z} \).
 With such observations, we aim to estimate the regression function
\[
\mathbb{E}[y | z] = m^*(z)
\]
using i.i.d. observations $(x_1, y_1), \ldots, (x_n, y_n)$, and
\begin{equation}
y_i = m^*(z_i) + \varepsilon_i, \quad \mathbb{E}[\varepsilon_i | z_i] = 0 
\label{eq:FAST}
\end{equation}
with i.i.d. noises $\varepsilon_1, \ldots, \varepsilon_n$. We aim to minimize the population $L_2$ error
$
\int ( \hat{m}(z) - m^*(z) )^2 \, \mu(dz)
$ with a focus on the setting that (1) $p$ is relatively high dimensional, and (2) the latent factor dimension $k$ is small.

In addressing this challenge, our approach involves introducing factor and additive structures to neural networks. This innovative methodology is designed to unveil the underlying non-linearity inherent in the data, providing a more comprehensive and interpretable representation of the complex relationships within the high-dimensional space.

\subsection{Our Contributions}
This paper aims to integrate factor model and additive model structures into neural networks for nonparametric regression modeling, with a focus on learning sparse \textbf{compositional structures} in a relatively high-dimensional regime. Our contribution can be summarized as follows:
\smallskip 
\begin{itemize}
    \item We explore Principal Component Analysis (PCA) as a method for factor estimation. In particular, we introduce a \textbf{differentiable PCA layer} that can be seamlessly integrated at any stage of a neural network model. To address the challenge of unstable factor axes, we have developed a training strategy that activates the PCA operation based on a predetermined operating schedule. This approach significantly improves the stability, performance, and efficiency of the PCA layers.

    \item We also introduce a \textbf{Soft PCA} layer designed to retain the variance of the output matrix while preserving a degree of orthogonality. This method integrates smoothly with the training process and does not require a predefined operating schedule.
    
    \item Leveraging the newly introduced factor layers, we propose the Generalized Factor Neural Network Model. Specifically, we present the Generalized Factor Augmented Neural Network (GFANN) and the Generalized Factor Additive Neural Network (GFADNN) architectures, which incorporate both factor layers and additive layers. This design enables neural networks to more effectively process data characterized by hierarchical composition models, as described in \citet{schmidt2020nonparametric}, particularly in high-dimensional settings.

    \item Our proposed Generalized Factor Neural Network Models demonstrate improved performance compared to the FAR-NN and FAST-NN across both simulated and real-world datasets. This advantage is particularly notable in scenarios involving non-linear relationships between latent factors and observed data. Our model's superior capability in capturing and modeling these complexities underscores its potential for broad application and further innovation in neural network design. 
    
\end{itemize}

\subsection{Notation}
We use bold lower case letter $\boldsymbol{x}=\left(x_1, \ldots, x_d\right)^{\top}$ to represent a $d$-dimension vector, let $\|\boldsymbol{x}\|_q=$ $\left(\sum_{i=1}^d\left|x_i\right|^q\right)^{1 / q}$ be its $\ell_q$ norm, and let $\|x\|_{\infty}=\max _{1 \leq i \leq d}\left|x_i\right|$ be its $\ell_{\infty}$ norm. We use bold upper case $\boldsymbol{A}=\left[A_{i, j}\right]_{i \in[n], j \in[m]}$ to denote a matrix. We define $\|\boldsymbol{A}\|=\sup _{\boldsymbol{x} \in \mathbb{R}^m,\|\boldsymbol{x}\|_2=1}\|\boldsymbol{A} \boldsymbol{x}\|_2$, let $\|\boldsymbol{A}\|_F=$ $\sqrt{\sum_{i, j} A_{i, j}^2}$, and let $\|\boldsymbol{A}\|_{\max }=\max _{i \in[n], j \in[m]}\left|A_{i, j}\right|$. Moreover, we use $\lambda_{\min }(\boldsymbol{A})$ and $\lambda_{\max }(\boldsymbol{A})$ to denote its minimum and maximum eigenvalue respectively. Please refer to table \ref{table:notation_acronyms} in Section \ref{sec:table_notation} for the table of notations and acronyms.

\subsection{Paper structure}
In Section~\ref{sec:section2}, we introduce the foundational models considered in this study. Section~\ref{sec:section3} details the integration of structures derived from factor and additive models into neural networks, leading to the development of PCA, Soft PCA, and Additive Layers as fundamental components. This section also provides the implementation algorithms. Section~\ref{subsec:Model_Architecture} and Section~\ref{subsec:GFADNN} elucidate the rationale behind the construction of our Generalized Factor Neural Network Model. Section~\ref{sec:section4} explores simulation studies and discusses methodologies for conducting fair model comparisons within the inherently stochastic nature of neural network outcomes, even with the same data input. Finally, Section~\ref{sec:section5} presents an empirical study focused on forecasting S\&P 500 ETFs and macroeconomic data sets, demonstrating the practical application of our proposed model.

\section{Model}
\label{sec:section2}
\subsection{Additive Models and Factor Models}

In financial studies, the returns of asset \( X_{j, i} \), for a given asset \( j \), over period \( i \), are influenced by shared factors. \citet{fan2021robust} employ the following model  
\begin{equation}
X_{j, i} = \sum_{m=1}^{d} b_{j, m} Z_{m, i} + u_{j, i},
\end{equation}
where \( j \) ranging from 1 to \( J \) stands for the total number of assets in the portfolio—and \( i \) stands for the portfolio's time horizon from 1 to \( n \). Here, \( d \) is the count of factors that are measured, with \( Z_{m, i} \) signifying the \( m \)-th factor at time \( i \), and \( b_{j, m} \) representing the factor loading for asset \( j \) and factor \( m \). The model specification imposes that the relation between the common factors and returns are linear. By incorporating the additive model mentioned in Section~\ref{subsec:nonpara}, we can allow a more flexible relationship between the asset returns and factors, we assume that the returns evolve through the following model
\begin{equation}
X_{j, i} = f_{j}(Z_i) + u_{j, i},
\end{equation}
where
\begin{equation}
f_{j}(Z_i) := \sum_{m=1}^{d} f_{j, m}(Z_{m, i}),
\end{equation}
and \( f_{j} \) denotes the true but unobservable function that links the factor matrix \( Z_i = (Z_{1, i}, \ldots, Z_{m, i}, \ldots, Z_{d, i})^\prime \), a \( d \times 1 \) vector, with each asset's returns. The function \( f_{j, m} \) is the unknown function for the \( m \)-th factor during period \( i \). Different assets can link to common factors through different functional forms. Compared to the strict conventional linear factor models, this model specification increases the flexibility in the relationship between the factors and the assets.


\subsection{ReLU Neural Networks}

We construct our model using a fully-connected deep neural network with ReLU activation $\sigma(\cdot) = \max\{\cdot, 0\}$, chosen for its empirical success. We refer to this as a deep ReLU network for brevity. Let $L \in \mathbb{N}$ be the depth of the network and $\boldsymbol{d} = (d_1, \ldots, d_{L+1}) \in \mathbb{N}^{L+1}$ define the dimensions of each layer. Following \citet{fan2023factor}'s definition, a deep ReLU network is a function mapping $\mathbb{R}^{d_0} \to \mathbb{R}^{d_{L+1}}$, expressed as
\begin{equation}
g(x) = \mathcal{L}_{L+1} \circ \bar{\sigma} \circ \mathcal{L}_L \circ \bar{\sigma} \circ \cdots \circ \mathcal{L}_2 \circ \bar{\sigma} \circ \mathcal{L}_1(x),
\label{eq:reluNN}
\end{equation}
where $\mathcal{L}_{\ell}(\boldsymbol{z}) = \boldsymbol{W}_{\ell} \boldsymbol{z} + \boldsymbol{b}_{\ell}$ represents an affine transformation with weight matrix $\boldsymbol{W}_{\ell} \in \mathbb{R}^{d_{\ell} \times d_{\ell-1}}$ and bias vector $\boldsymbol{b}_{\ell} \in \mathbb{R}^{d_{\ell}}$. The operator $\bar{\sigma}: \mathbb{R}^{d_{\ell}} \to \mathbb{R}^{d_{\ell}}$ applies the ReLU activation function elementwise to a $d_{\ell}$-dimensional vector. For simplicity, we refer to both $\boldsymbol{W}_{\ell}$ and $\boldsymbol{b}_{\ell}$ as the network's weights.

The family of deep ReLU networks, truncated at level $M$ with depth $L$, width parameter $\boldsymbol{d}$, and weights bounded by $B$, is defined as
\begin{equation}
\mathcal{G}(L, \boldsymbol{d}, M, B) = \left\{\widetilde{g}(\boldsymbol{x}) = \bar{T}_M(g(\boldsymbol{x})) : g \text{ of form \eqref{eq:reluNN} with } 
\|\boldsymbol{W}_{\ell}\|_{\max} \leq B, \|\boldsymbol{b}_{\ell}\|_{\max} \leq B\right\},
\end{equation}
where $\bar{T}_M(\cdot)$ is the truncation operator at level $M$ applied element-wise to a $d_{L+1}$-dimensional vector, more specifically defined as 
\[
\left[\bar{T}_M(z)\right]_i = \operatorname{sgn}(z_i)\min\{|z_i|, M\}.
\]
If the width parameter is given as $\boldsymbol{d} = (d_{\text{in}}, N, N, \ldots, N, d_{\text{out}})$, we denote it as $\mathcal{G}(L, d_{\text{in}}, d_{\text{out}}, N, M, B)$, referring to a deep ReLU network with depth $L$ and width $N$ for brevity.

For practical implementation, LeakyReLU is selected as the activation function due to its ability to mitigate the 'dying ReLU' issue \citet{maas2013rectifier}. By permitting a minor gradient even when the neuron is inactive, LeakyReLU ensures the continuity of gradient flow during the entire training cycle. This characteristic not only potentially accelerates the convergence rate, but also fosters more uniform learning across deep neural networks.

\section{Methodology}
\label{sec:section3}
In this section, we delineate the methodology underpinning our model for high-dimensional regression, focusing on two complementary approaches. First, we incorporate factor structure into the neural network using the standard PCA Layer and the novel Soft PCA Layer. Second, we introduce an additive structure through the Additive Layer to impose a low-dimensional structure on the regression function.

\subsection{Differentiable PCA Layer}
\begin{figure}[H]
\centering
    \includegraphics[width=\columnwidth]{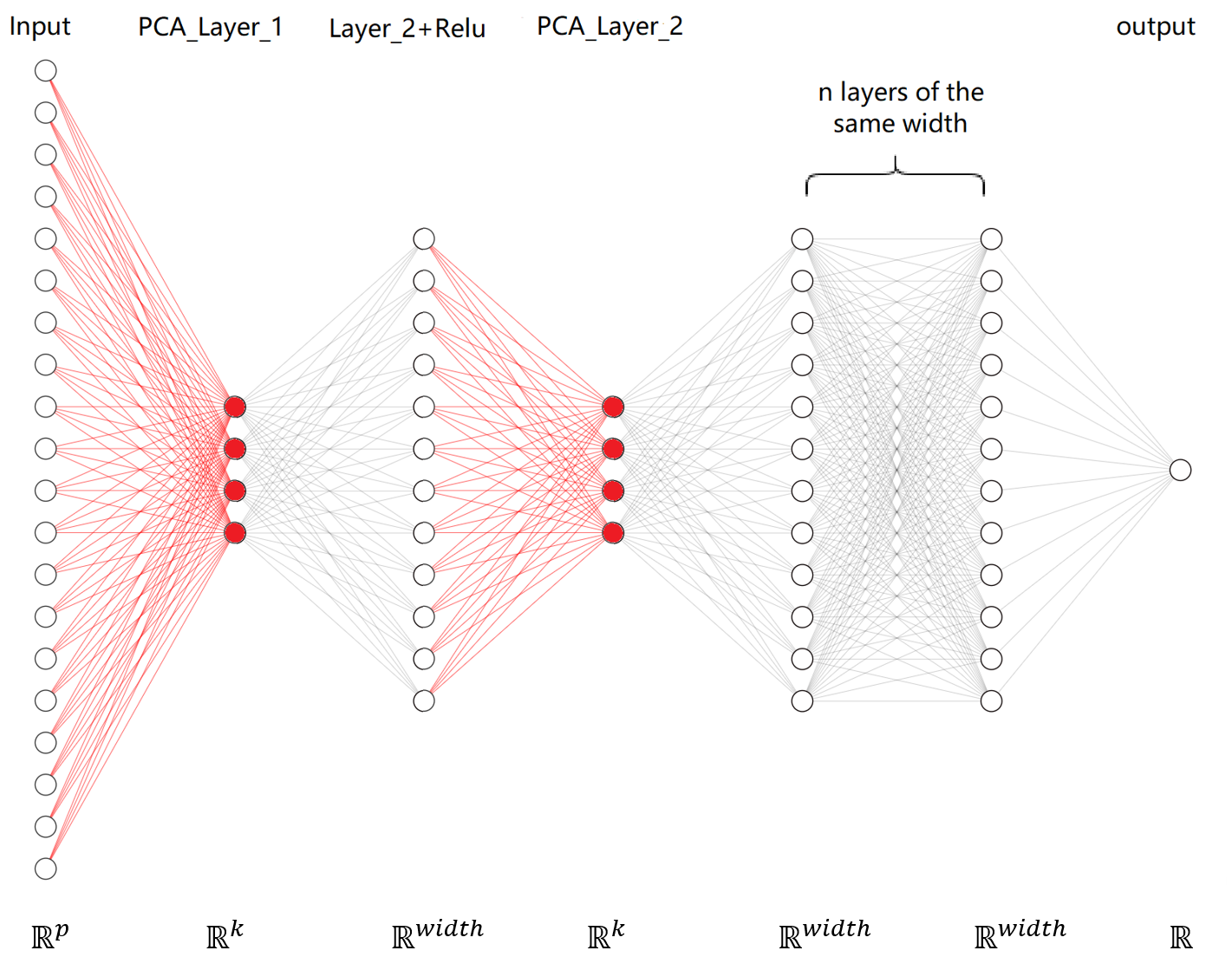}
    \caption{A visual illustration of integrating multiple PCA layers into neural networks. The layers in red denote the PCA layers. }  
\label{fig:pca_pca_nn}
\end{figure}

The work by \Citet{fan2023factor} introduced a model incorporating a diversified projection matrix prior to the neural network layer, estimated through Principal Component Analysis (PCA). However, their approach restricts PCA to the initial step of the model. When the data does not follow a linear factor model or exhibits a hierarchical factor structure, alternating between factor modeling and non-linear transformations becomes essential \citep{bengio2013representation}. Even when the data follows a linear factor structure, performing PCA multiple times at intermediate layers, as illustrated in Figure \ref{fig:pca_pca_nn}, can help refine feature representations, reduce redundancy, and mitigate noise accumulation, thereby improving training stability and representational efficiency \citep{goodfellow2016deep}. This necessitates the ability to insert the PCA operation at any stage of a neural network architecture, enhancing its adaptability and effectiveness in high-dimensional settings \citep{fan2014challenges}.

A significant challenge of this approach lies in ensuring the differentiability of the PCA operation while maintaining end-to-end training stability. This can be difficult to achieve if the PCA operation causes drastic changes to weight parameters after processing each batch of data. To address this challenge, we introduce methods to stabilize and seamlessly integrate PCA within the training process. Specifically, we implement PCA operations using differentiable eigenvector computations. The accompanying pseudo-code is provided in Algorithm 1, serving as a procedural guide. For more intricate technicalities, we refer to \textit{On differentiating eigenvalues and eigenvectors} by \citet{magnus1985differentiating}. It is imperative to note that the gradient can become numerically unstable when the disparity between any two eigenvalues approaches nullity, as it is contingent upon the eigenvalues \( \lambda \) through the computation of \( \frac{1}{\min_{i \neq j} (\lambda_i - \lambda_j)} \). To circumvent this, our algorithm instates a stability criterion that ascertains whether the minimal difference among the eigenvalues surpasses a predefined threshold. Should this not be the case, we introduce a perturbation by adding a scalar multiple of random values, scaled according to the magnitude of diagonal values, to the covariance matrix to enhance stability.

\begin{algorithm}[H]
   \caption{PCA Layer}
   \label{alg:Factor Layer}
\begin{algorithmic}[1]  
   \State {\bfseries Input:} data $A$, shape $n \times p$
   \State {\bfseries Output:} data $B$, shape $n \times k$
   \State Initialize placeholder $C$, shape $p \times k$
   \If{$\textit{initializing}$}
      \State $covMatrix \gets \text{cov}(A^T)$
      \State $eigenvalues, eigenvectors \gets \text{eigen}(covMatrix)$
      \Comment{Stability check}
      \While{$\min(eigenvalues.\text{diff}()) \leq 10^{-10}$}
         \State \# add noise to the cov matrix
         \State $covMatrix \mathrel{{+}{=}} 10^{-4} \times \text{mean}(\text{diag}(covMatrix)) \times \text{randLike}(covMatrix)$
         \State $eigenvalues, eigenvectors \gets \text{eigen}(covMatrix)$
      \EndWhile
      \State $C \gets eigenvectors[:,:k] / sqrt(p)$
      \State $factor \gets A \cdot C$
      
      \State \textbf{return} $factor$
   \Else
      \State $factor \gets A \cdot C$
      \State \textbf{return} $factor$
   \EndIf
\end{algorithmic}
\end{algorithm}

\subsubsection{Unstable Factor Axes}
However, when training a neural network model with multiple PCA layers as shown in Figure 1, we observed a markedly higher test error relative to vanilla neural networks. This phenomenon appears rooted in the instability of factor axes. Conceptually, PCA is akin to fitting a p-dimensional ellipsoid around the data, with each principal component acting as an ellipsoid axis. These axes, set sequentially based on the direction of maximum variance in projections, are susceptible to alteration with varying data batches and evolving input during training. Figure 3 illustrates the training progression within one of our simulated data experiments. It reveals notable instability in both training and validation error, with the test error also exhibiting inferior performance compared to that of a standard neural network.

\begin{figure}[H]
\centering
    \includegraphics[width=10cm ]{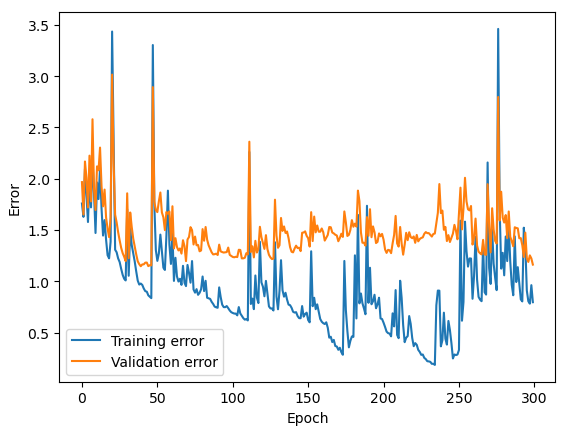}
    \caption{Training and validation error when training a Neural Network model with a PCA layer in the middle.}
\end{figure}

\begin{figure}[H]
\centering
    \includegraphics[width=10cm ]{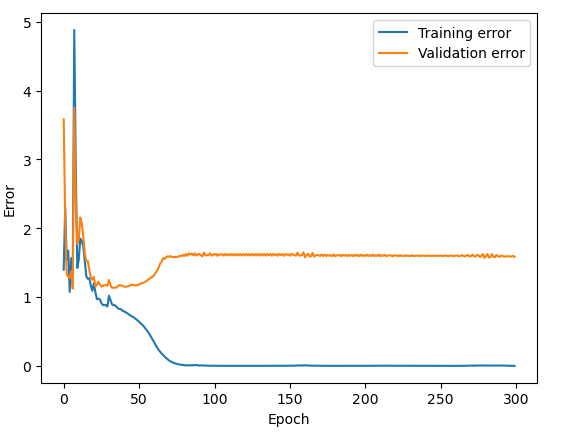}
    \caption{Training and validation error when training a Neural Network model with a PCA layer in the middle, with an operating schedule.}
    \label{fig:train_valid_loss_schedule}
\end{figure}

To address this instability, we propose an operating schedule for the model. This approach enables PCA decomposition exclusively during initialization (i.e., when $initializing==True$). This allows us to reduce the number of PCA operations and hence reduce noise introduced by the unstable factor axes during training. Intuitively, with each PCA update, the model requires a certain number of epochs to align its weights. As the model attains stability, the need for frequent PCA recalibrations diminishes. Consequently, a tailored operating schedule that incrementally decreases the PCA operations ensures stable factor axes while accommodating model evolution. Figure \ref{fig:train_valid_loss_schedule} presents the model's training process with an operating schedule, applying PCA operations at the initial batch of each specified epoch—namely epochs [1, 3, 5, 7, 9, 11, 13, 15, 20, 30, 40]. A rapid stabilization of both training and validation errors can be observed. The refinement not only enhances test results but also markedly decreases computational complexity.

\subsubsection{Monitoring Explained Variance}
\begin{figure}[H]
\centering
    \includegraphics[width=10cm ]{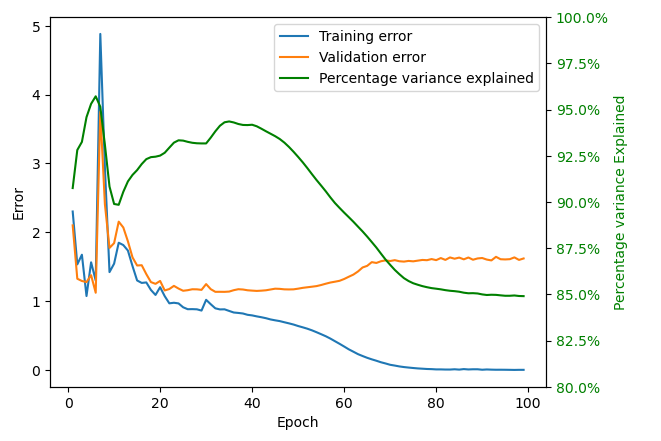}
    \caption{Training and Validation error with an operating schedule of [1,2,...40] and the corresponding percentage variance explained by the first $k$ components}
    \label{fig:monitorvariance}
\end{figure}

In typical applications, a PCA layer serves to reduce the dimensionality of the input data matrix from $p$ dimensions to $k$. It also enables us to monitor the variance explained throughout the training process. Figure~\ref{fig:monitorvariance} depicts the training process of the neural network model, with an operating schedule spanning epochs 1 through 40. The green line illustrates the percentage of variance explained by the first $k$ principal components. A discernible increasing trend in the percentage of variance explained is observed, and then it stabilizes, coinciding with the stabilization of the validation error. This trend not only sheds light on the sufficiency of the chosen width $k$ in capturing the majority of the variance, but also offers insights into the linear interpretability of the input matrix.

\subsubsection{Limitations and Challenges}
Non-adaptive PCA layers are inherently rigid, requiring precomputed transformations or periodic updates based on fixed schedules. This rigidity becomes particularly limiting in dynamic or nonstationary data environments, where the statistical properties of the data evolve over time. Moreover, PCA-based factor estimation is highly sensitive to eigenvalue distributions; when eigenvalues are close in magnitude, small perturbations in the data can lead to instability. While perturbation methods can mitigate this issue, they introduce additional computational complexity. Additionally, freezing factor weights after initial training preserves exogeneity but limits adaptability when the data distribution shifts

\subsection{Soft PCA Layer}
In this subsection, we introduce the \textbf{Soft PCA Layer}, a novel neural network layer designed to approximate the effects of PCA without relying on a predefined operating schedule. Unlike traditional PCA, which relies on deterministic eigendecomposition, the Soft PCA Layer integrates PCA-like behavior into the learning process, enabling it to adapt dynamically during training. The Soft PCA Layer addresses above-mentioned challenges by embedding PCA-inspired objectives into a fully trainable framework, allowing it to:
\begin{itemize}
    \item Adapt dynamically to evolving data representations across network layers.
    \item Eliminate the need for predefined update schedules, reducing manual intervention.
    \item Achieve PCA-like variance maximization and orthogonality without explicit eigendecomposition.
\end{itemize}

\subsubsection{Design and Objectives}
The Soft PCA Layer achieves its objectives through two key components:

\textit{1. Variance Maximization Objective}: Traditional PCA maximizes variance along principal components. To approximate this behavior, we minimize the loss function
\[
\mathcal{L}_{\text{variance}} = \| \text{Var}(\mathbf{X}_{\text{in}}) - \text{Var}(\mathbf{X}_{\text{out}}) \|_F^2,
\]
where \( \mathbf{X}_{\text{in}} \) and \( \mathbf{X}_{\text{out}} \) represent the input and output of the layer, respectively, and \( \| \cdot \|_F \) is the Frobenius norm. This ensures that the transformation captures the most significant variations in the input.

\textit{2. Orthogonality Constraint}: To encourage outputs resembling the orthogonal projections of PCA, we impose a loss term on the covariance matrix \( \Sigma_{\text{out}} \) of the layer outputs
\[
\mathcal{L}_{\text{orthogonality}} = \| \Sigma_{\text{out}} - \text{diag}(\Sigma_{\text{out}}) \|_F^2,
\]
where \( \text{diag}(\Sigma_{\text{out}}) \) is the diagonal of the covariance matrix. By minimizing off-diagonal elements, the layer approximates the orthogonality of PCA's principal components.

\subsubsection{Integration into Neural Networks}
The Soft PCA Layer is implemented as a fully differentiable layer, making it compatible with standard backpropagation. Unlike the traditional PCA preprocessing step, it operates as a dynamic and trainable transformation embedded within the neural network. This flexibility enables its integration at any stage of the model, much like the standard PCA Layer described earlier.

\subsubsection{Comparison with Standard PCA Layer and Autoencoders}
While the standard PCA Layer provides deterministic dimensionality reduction and variance decomposition, it requires periodic updates or fixed preprocessing. In contrast, the Soft PCA Layer offers
\begin{itemize}
    \item \textbf{Adaptability}: It adapts to evolving feature representations during training.
    \item \textbf{End-to-End Learnability}: Integrated into the model, it eliminates the need for separate preprocessing.
    \item \textbf{Computational Efficiency}: Avoids the computational cost of explicit eigendecomposition during training.
\end{itemize}

In comparison to autoencoders, the Soft PCA Layer excels in scenarios involving high-dimensional data with low-rank structure due to its built-in orthogonality constraints and variance maximization objective. These properties make the Soft PCA Layer particularly suitable for extracting uncorrelated features and preserving the most significant variations in the data, aligning closely with the goals of PCA. While autoencoders are powerful for nonlinear feature learning, they introduce additional parameters and lack explicit decorrelation objectives, making them less ideal for problems where interpretability and efficient handling of low-rank structures are critical.

The flexibility and adaptability of the Soft PCA Layer make it particularly suitable for scenarios where data characteristics or feature representations evolve, such as in high-dimensional regression tasks or dynamic systems modeling.



\subsection{Factor Estimation via Diversified Projection Matrix}
We introduce the idea of a \textit{diversified projection matrix} proposed by \citet{fan2022learning}, in order to provide more intuition for the frozen factor layer. 
\begin{definition}[Diversified projection matrix]
Let $\bar{k} \ge k$, and $c_1$ be a universal positive constant. A $p \times \bar{k}$ matrix $\mathbf{W}$ is said to be a diversified projection matrix if it satisfies
\begin{enumerate}
    \item \textbf{(Boundedness)} $\|\mathbf{W}\|_{\max} \le c_1$.
    \item \textbf{(Exogeneity)} $\mathbf{W}$ is independent of $x_1, \dots, x_n$ in \eqref{eq:non-linear_factor}.
    \item \textbf{(Significance)} The matrix $\mathbf{H} = p^{-1} \mathbf{W}^\top \mathbf{B} \in \mathbb{R}^{\bar{k} \times k}$ satisfies $\nu_{\min}(\mathbf{H}) \gg p^{-1/2}$.
\end{enumerate}
Each column of \( \mathbf{W} \) is referred to as a diversified weight, with \( \bar{k} \) representing the total number of diversified weights.
\end{definition}

The core concept behind the \textit{diversified projection matrix} is that we can construe
\[
\tilde{\mathbf{z}} = p^{-1} \mathbf{W}^\top \mathbf{x}
\]
as a proxy for the factor \( \mathbf{z} \) in downstream predictions, even when \( \bar{k} > k \), effectively overestimating the number of factors. To see why this works, we can substitute equation \eqref{eq:non-linear_factor} into the expression above, yielding
\begin{equation}
\tilde{\mathbf{z}} = \mathbf{H} g(\mathbf{z}) + \bm{\xi}, \quad \text{where } \bm{\xi} = p^{-1} \mathbf{W}^\top \mathbf{u}.
\label{eq:decompose_proj}
\end{equation}

This decomposition shows that, under mild conditions, \( \tilde{\mathbf{z}} \) provides a good estimate of an affine transformation of \( g(\mathbf{z}) \). The intuition here is that if the idiosyncratic component \( \mathbf{u} \) has weak dependence and uniformly bounded second moments, then \( \|\bm{\xi}\| = O_P(p^{-1/2}) \) due to the bounded variance. Meanwhile, due to the \textit{significance} condition and the presence of non-degenerate factors, the signal term satisfies
\[
\| \mathbf{H} g(\mathbf{z}) \|_2 \ge \nu_{\min}(\mathbf{H}) \| \mathbf{z} \|_2 \gg \| \bm{\xi} \|_2,
\]
implying that \( \mathbf{H} g(\mathbf{z}) \) is the dominant term in the decomposition \eqref{eq:decompose_proj}.

We can intuitively interpret the diversified projection matrix \( \mathbf{W} \) as an ``overestimate" of the factor loading matrix \( \mathbf{B} \). The term ``over-" indicates that it is not necessary to precisely determine the number of factors \( k \) in our framework; instead, we can simply choose a sufficiently large \( \bar{k} \). 

Moreover, the \textit{significance}  condition requires that the choice of \( \mathbf{W} \) be more informed than a random guess. For instance, for an \( n \times n \) random matrix \( \mathbf{X} \) with i.i.d. Gaussian entries, the smallest singular value \( \sigma_{\min}(\mathbf{X}) \) scales as \( O(n^{-1/2}) \). This does not satisfy the \textit{significance} condition for a diversified projection matrix. 

In the case of our (Soft) PCA layer, one may clip the weights to ensure we satisy the \textit{boundedness}  condition, although, in practice, the initial weight values also generally meet this requirement. Regarding the \textit{exogeneity} condition, \citet{fan2023factor} documented that when the weights are trained jointly with other layers (and thus become dependent on \( x_1, \dots, x_n \)), performance deteriorates significantly. A potential approach to maintain exogeneity is to estimate \( \mathbf{W} \) using an independent set of observations, separate from \( x_1, \dots, x_n \). However, in practice, we opt to freeze the weights after a few epochs, which prevents them from further adapting to \( x_1, \dots, x_n \).

Lastly, the (Soft) PCA layer satisfies the \textit{significance} condition, as it is either orthogonal due to the PCA operation or approximately orthogonal in the Soft PCA case. When a matrix has approximately orthogonal columns, its singular values are generally well-distributed and bounded away from zero. Additionally, because the layer approximates the projection from \( \mathbf{x} \) to \( \mathbf{g(z)} \), it aligns with \( \mathbf{B} \) to a certain extent after the initial training stage. This alignment prevents any particular direction from collapsing when multiplied by \( \mathbf{B} \), thereby preserving the rank and significance of \( \mathbf{H} \).

\subsection{Additive Layer}
In the next phase of our pipeline, we focus on integrating an additive structure into the neural network. This approach involves isolating specific sections of the previous layer to form sub-networks and combining them solely through addition. Consider a scenario where we have identified latent factors $X$ with dimensions $n \times p$ following a PCA layer. 

Rather than forwarding this output to subsequent fully connected layers or the final output layer, we apply the additive structure as shown in Algorithm \ref{alg: Additive_Layer}: we divide $X$ column-wise into segments $[X_{p_1}, X_{p_2}, ..., X_{p_j}]$. Each segment $g(X)_{p_i}$ is characterized by the dimensions $n \times p_i$, ensuring that the sum of all $p_i$ equals $p$ (i.e., $\sum^j_i p_i = p$). Each sub-matrix $X_{p_i}$ is then fed into a linear layer of size $p_i \times q_i$. Consequently, the output from each linear layer retains the shape $n \times q_i$. Then we horizontally concatenate the output matrices, resulting in a matrix $[X_{q_1}, X_{q_2}, ..., X_{q_j}]$ of shape $n \times q$, where $\sum^j_i q_i = q$. When constructing the layer,we need to specify the $input\_dimension\_list\ [p_1, p_2..p_i]$ and  $output\_dimension\_list\ [q_1, q_2..q_j]$ By aligning the $input\_dimension\_list$ of a subsequent Additive Layer with the $output\_dimension\_list$ of the preceding one, (i.e. the $input\_dimension\_list$ of the subsequent Additive Layer is $[q_1, q_2..q_j]$) we ensure that there is no interaction among the initial column segments $[X_{p_1}, X_{p_2}, ..., X_{p_j}$. Through one or more such Additive Layers, our neural network can be effectively segmented into multiple sub-networks. In the final step, we aggregate these sub-networks, forming the additive structure. Figure \ref{fig:additive_layer} in Section \ref{sec:additive_layer_graph} provides an illustration of the additive layers.


\begin{algorithm}[H]
   \caption{Additive Layer}
   \label{alg: Additive_Layer}
\begin{algorithmic}
   \State {\bfseries Input:} data $A$, shape $n \times p$; $input\_dim\_list = [p_1, p_2, \ldots]$ where $\sum^j_{i=1} p_i = p$
   \State {\bfseries Output:} data $B$, shape $n \times q$; $output\_dim\_list = [q_1, q_2, \ldots]$ where $\sum^j_{i=1} q_i = q$
   \State $n\_subNN \gets \text{len}(input\_dim\_list)$
   \State $subNN\_idx \gets [0] \cup \text{cumsum}(input\_dim\_list)$ \Comment{Indices to partition $A$}
    \State $layer\_list \gets [$
    $\text{linear\_layer}(input\_dim\_list[I], output\_dim\_list[i])$
    \State \hspace{2.5cm} $\text{for} \ i \ \text{in} \ \text{range}(n\_subNN)]$
    
    \State $output\_list \gets [$
    $layer\_list[i](A[:,$
    $subNN\_idx[i]:subNN\_idx[i+1]])$
    \State \hspace{2.5cm} $\text{for} \ i \ \text{in} \ \text{range}(n\_subNN)]$
   \State $B \gets \text{concat}(output\_list)$
   \State \textbf{return} $B$
\end{algorithmic}
\end{algorithm}

\subsection{Model Architecture}
\label{subsec:Model_Architecture}

We can now present some intuition on how to design the model architectures based on the data-generating process with an underlying low-dimensional structure. Assuming that we have observable $n\times 1$ target $Y$, and high-dimensional $n\times p$ features $X$ and there exist $n\times k$ latent factors $Z$ where $k << p$. In a linear setting, the observations have a factor structure in a way that $X=BZ+\epsilon$, where $B$ is the loading matrix and $\epsilon$ is the noise term. We specify the target $y=f(Z)+\mu$ to follow an additive structure: $y_i=\sum_{j \in \mathcal{J}} f_j\left(z_{i, j}\right)+\varepsilon_i \quad$ for $\quad i=1, \ldots, n$. In this data-generating process, the latent factors $Z$ bridges observations $X$ and the target $Y$ with $Z-X$ structure and $Z-Y$ structure. We can further add some non-linearity to the generating process by passing $Z$ through some simple non-linear functions $g$ like $B \times g_1\left(\left[z_1, z_2, z_3\right]\right)$. 
Under this more general setting, our model needs to pass the input through some layers to approximate $g$ and then pass to a Factor Layer to perform factor decomposition. Next, since we assume $Z-Y$ structure to be additive, we need to pass the data to multiple Additive Layers so that each subnetwork formed can approximate the non-linear $f_i$ function. Following this intuition, we can further model higher-order/hierarchical additive factor models.\\\\

\subsubsection{Generalized Factor Additive Neural Network Model}  
\label{subsec:GFADNN}

\begin{figure}[H]
\centering
\includegraphics[width=\columnwidth]{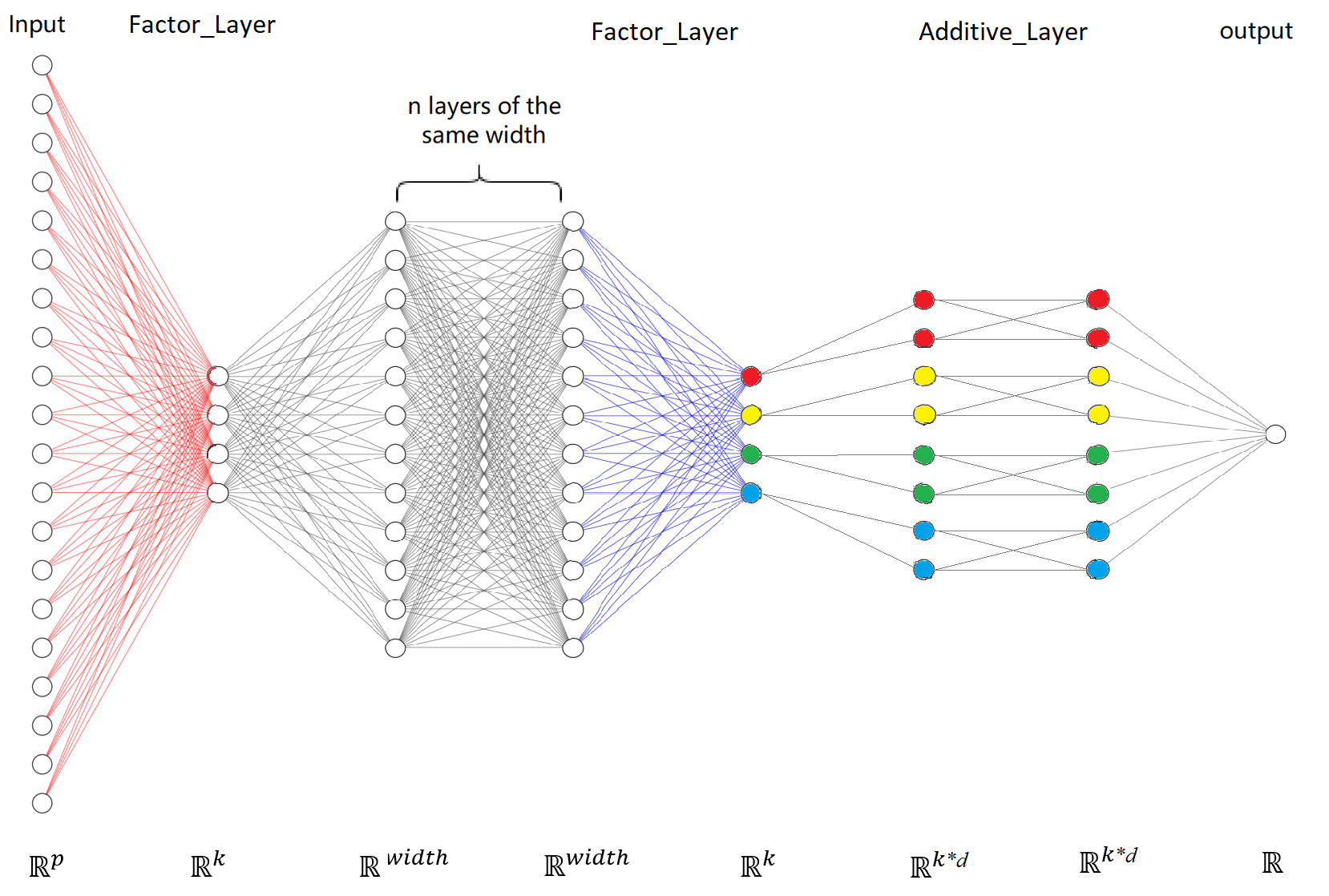}
\vspace{1cm} 

\caption{An illustration of the General Factor Additive Neural Network Model. The red and blue layers represent generalized factor layers, while the final layers are additive layers. The input to the additive layers is already (partially) decorrelated before being passed through them.}
\label{fig:FADNN}
\end{figure}

The rationale for integrating fully-connected layers between the PCA layer and the Additive layers stems from the need to address potential non-linearity between the latent factors and the observations. However, considering that a primary motivation for employing Additive layers is to segregate information flows, the introduction of fully-connected layers might counteract this by amalgamating the orthogonal outputs derived from the PCA layers. Therefore, to preserve the discrete transfer of information to the Additive layers, it is preferable to limit neuron interactions, which can be achieved by directly connecting the Additive layers to the PCA layer.


\subsubsection{Generalized Factor Augmented Neural Network Model}  
\label{subsec:GFANN}
Following FAST-NN estimator, we propose the Generalized  Factor Augmented Neural Network (GFANN) Model. Recall the non-linear factor model specification in equation \eqref{eq:non-linear_factor}. With this latent factor structure, it is natural to consider a factor-augmented regression model where $z$ and $u$ serve as regressors. This approach is equivalent to using both $x$ and latent factors $z$ as regressors, but the former formulation results in weaker dependence among variables. We can then formulate the regression as such:
\[
\mathbb{E}[y | z, \mathbf{u}] = m^*(z, u_{\mathcal{J}}),
\]
where $\mathcal{J} \subset \{1, 2, \ldots, p\}$ is an unknown subset of indexes.In the GFANN model, the input matrix \( \mathbf{X} \in \mathbb{R}^{n \times p} \) is first passed through a layer that acts as a projection matrix \( \mathbf{P} \in \mathbb{R}^{p \times k} \) to produce the output \( o_1 \). Next, the residual matrix \( \mathbf{u} \in \mathbb{R}^{n \times p} \) is computed as \( \mathbf{u} = \mathbf{X} - \mathbf{P} \mathbf{P}^T \mathbf{X} \). This residual is then passed through a variable selection matrix \( \mathbf{Q} \in \mathbb{R}^{p \times m} \) to obtain the output \( o_2 \). Finally, \( o_1 \) and \( o_2 \) are concatenated and passed to the subsequent network layers. The Figure \ref{fig:GFANN} below illustrates the network architecture.

\begin{figure}[H]
\centering
\includegraphics[width=1.2\columnwidth]{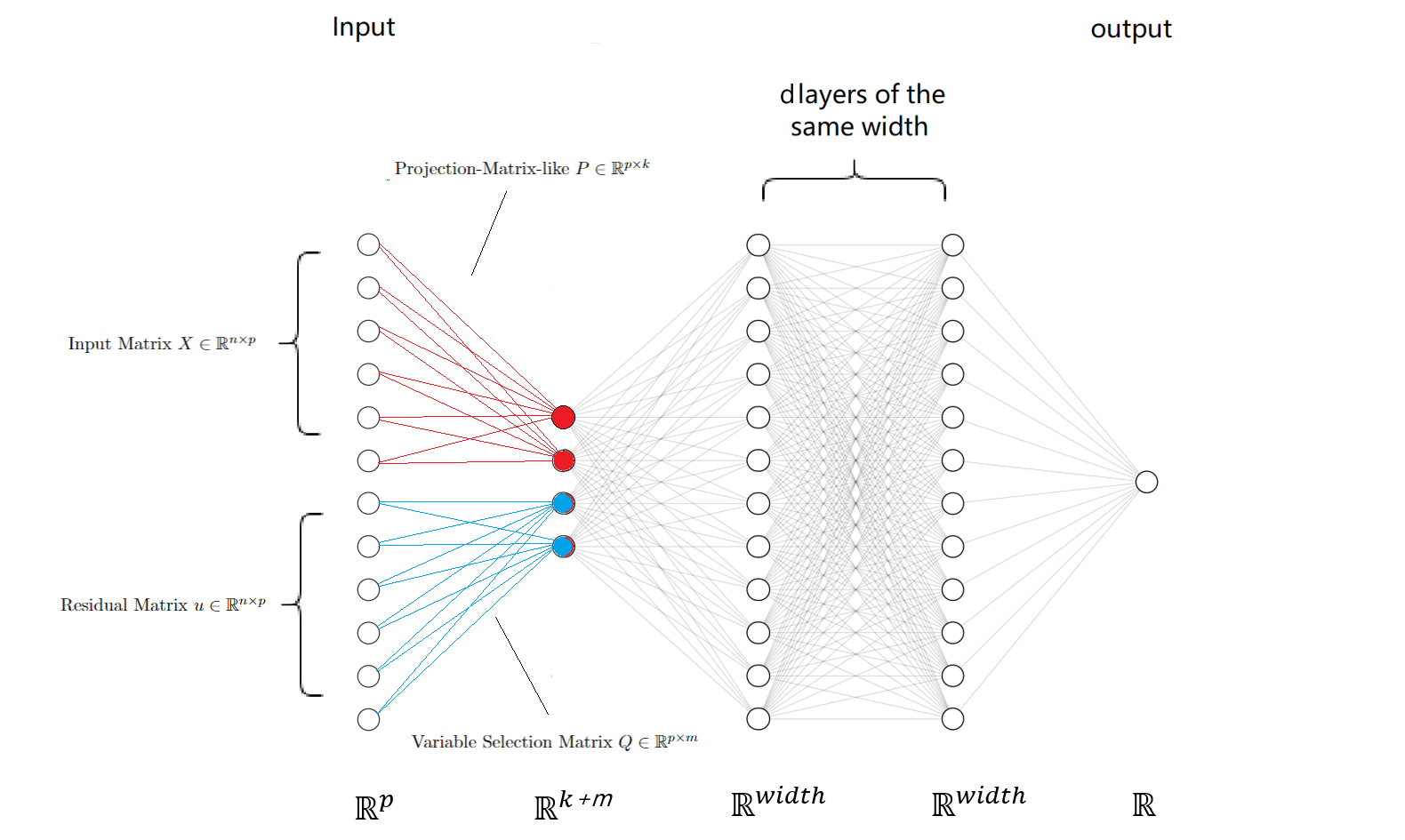}
\caption{Visualization of the Generalized Factor Augmented Neural Network Model. Layers in red and blue are factor layers}
\label{fig:GFANN}
\end{figure}

\newpage
\section{Simulation Studies}
\label{sec:section4}
\subsection{Data Generating Process}

\begin{table}[H]
\centering
\begin{tabular}{|l|l|l|} 
\hline
Observations (Z-X structure) & Target (Z-Y structure)\\
\hline
{1. $B \times \left[z_1, z_2, z_3\right]$} & 1. $ f_1\left(z_1\right)+f_2\left(z_2\right)+f_3\left(z_3\right)$\\
\hline
{2. $B \times g_1\left(\left[z_1, z_2, z_3\right]\right)$ }  &  1. $ f_1\left(z_1\right)+f_2\left(z_2\right)+f_3\left(z_3\right)$ \\
\hline
\end{tabular}
\label{tab:Variations of data structuress} 
\vspace{5pt}
\caption{Variations of data structures of observations and the target}
\end{table}

For the data-generating process, we follow the basic setup in \citet{fan2023factor}. For the $Z-X$ structure, we first generate $k=5$ independent latent factors, which follow Uniform$[-1,1]$, so the shape of the latent factor matrix $Z$ is $n$ by 5. The factor loading matrix $B$ has i.i.d. Unif $[-\sqrt{3}, \sqrt{3}]$ entries, and the shape is 5 by $p$. To generate observations, for case 1 (obs-id 1), we multiply $Z$ and $B$, then add idiosyncratic components $\boldsymbol{u}$, which are independent and have i.i.d. Unif $[-1,1]$ entries; for case 2 (obs-id 2), we pass $Z$ to a non-linear function $g$ first, where $g$ assumed to be some intuitive non-linear functions, in this case $exp$, since it is common and have economic intuition. To get the observation matrix $X$, we multiply $g(Z)$ by $B$ then add idiosyncratic components $\boldsymbol{u}$. To generate the target $Y$, in case 1 (target-id 1), we pass each latent factor $z_i$ to a non-linear function $f_i$, where $f_i$ are selected randomly from the candidate function set $\left\{\cos (\pi x), \sin (x),(1-|x|)^2, 1 /\left(1+e^{-x}\right), 2 \sqrt{|x|}-1, x^2\right\},$ in each trial. Finally, we add an independent noise term $\varepsilon$ which follows a zero-mean Gaussian distribution with a variance of $\sigma_\varepsilon=0.3$. For a high-dimensional setup, we will use $n_{\text {train }}=500$ i.i.d. samples from the above data-generating process to train our neural network. We also use other $n_{\text {valid }}=150$ i.i.d. observations as a validation data set for model selection and $n_{\text {valid }}=10000$ i.i.d. observations as a test data set.
\subsection{Model Comparison}
There is a substantial body of literature that compares neural network models using identical hyperparameter settings. However, hyperparameter settings are highly specific to each model, and therefore, for a fair comparison, we aim to select the best hyperparameter setting for each individual model under multiple random seeds. For training details, please refer to Section \ref{sec:model_comparison}. At this stage, the following models are selected for comparison: 
\begin{itemize}
    \item A vanilla Relu-Neural network model with \textbf{latent factors} as input (oracleNN).
    \smallskip 
    \item A vanilla Relu-Neural network model (vanillaNN).
    \smallskip 
    \item FAR-NN estimator \citep{fan2023factor} that motivated this research ($ FAR-NN$).
    \smallskip 
    \item A Generalized Factor Neural Network with Soft PCA as the first layer ($SPCA\_NN$) for the linear case and the second layer  ($NN\_SPCA\_NN$) for the non-linear case.
    \smallskip 
    \item A GFADNN model with PCA layers ($PCA\_NN\_PCA\_ADD$).
    \smallskip 
    \item A GFADNN model with soft PCA layers ($SPCA\_NN\_SPCA\_ADD$).
    \smallskip 
\end{itemize}
\subsection{Results}

\begin{table}[h!]
\centering
\begin{tabular}{|c|c|l|c|}
\hline
\textbf{obs-id} & \textbf{target-id} & \textbf{model} & \textbf{test mse} \\ \hline
1 & 1 & SPCA\_NN\_SPCA\_ADD & 0.038 \\ \hline
1 & 1 & PCA\_NN\_PCA\_ADD & 0.052 \\ \hline
1 & 1 & oracleNN & 0.080 \\ \hline
1 & 1 & FAR-NN & 0.089 \\ \hline
1 & 1 & SPCA\_NN & 0.093 \\ \hline
1 & 1 & autoencoder & 0.230 \\ \hline
1 & 1 & vanillaNN & 0.539 \\ \hline
1 & 1 & lasso & 1.507 \\ \hline
1 & 1 & pcr & 2.138 \\ \hline
2 & 1 & PCA\_NN\_PCA\_ADD & 0.065 \\ \hline
2 & 1 & SPCA\_NN\_SPCA\_ADD & 0.078 \\ \hline
2 & 1 & oracleNN & 0.080 \\ \hline
2 & 1 & FAR-NN & 0.103 \\ \hline
2 & 1 & NN\_SPCA\_NN & 0.151 \\ \hline
2 & 1 & autoencoder & 0.455 \\ \hline
2 & 1 & vanillaNN & 0.722 \\ \hline
2 & 1 & lasso & 1.444 \\ \hline
2 & 1 & pcr & 1.511 \\ \hline
\end{tabular}
\vspace{5pt}
\caption{Performance table with input dimension $p=500$, for two observation scenarios (obs-id 1 and 2).}
\label{table:test_mse_combined}
\end{table}

When the observations take the form \( B \times \left[z_1, z_2, z_3\right] \), $obs-id 1$, the relationship between the observations and latent variables is linear. Model performance is evaluated based on test mean squared error (MSE), with no noise term included in the test data set. Starting from an input dimension of \( p = 500 \), our GFADNN models consistently outperform the other models, with the \( SPCA\_NN\_SPCA\_ADD \) model improving the test MSE of \( FAR\_NN \) by 57.3\%.

When the observations take the form \( B \times \exp\left(\left[z_1, z_2, z_3\right]\right) \), $obs-id 2$, introducing non-linearity into the relationship between observations and latent variables, our GFADNN models again consistently outperform all others. In this case, the \( PCA\_NN\_PCA\_ADD \) model improves the test MSE of \( FAR-NN \) by 36.9\%.
\newpage

\subsection{Performance against input dimension}

\begin{figure}[H]
\centering

\begin{subfigure} 
    \centering
    \includegraphics[width=0.95\textwidth]{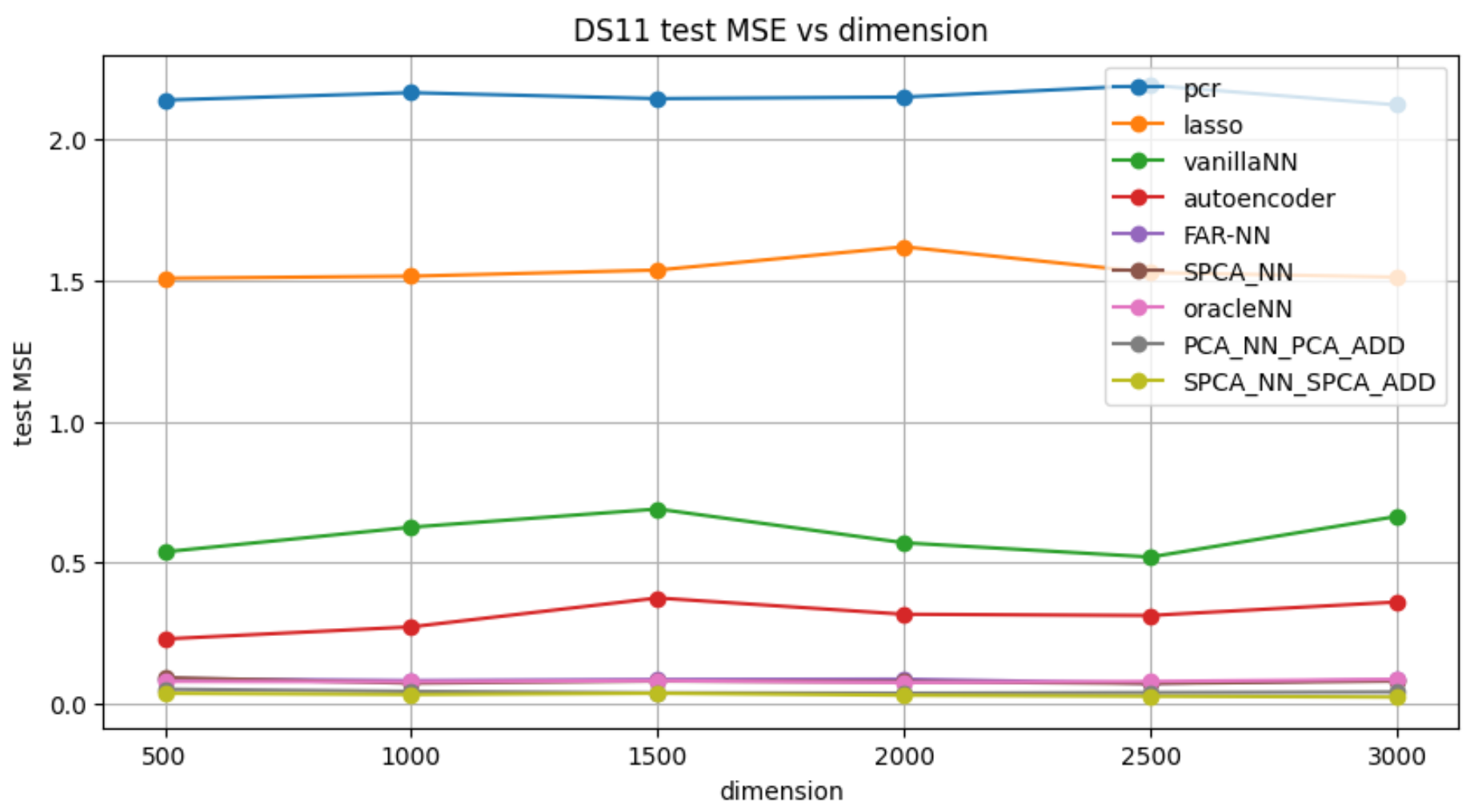}
    \caption*{(a)}
    \label{fig:ds02_table1}
\end{subfigure}
\hfill 

\begin{subfigure}
    \centering
    \includegraphics[width=0.95\textwidth]{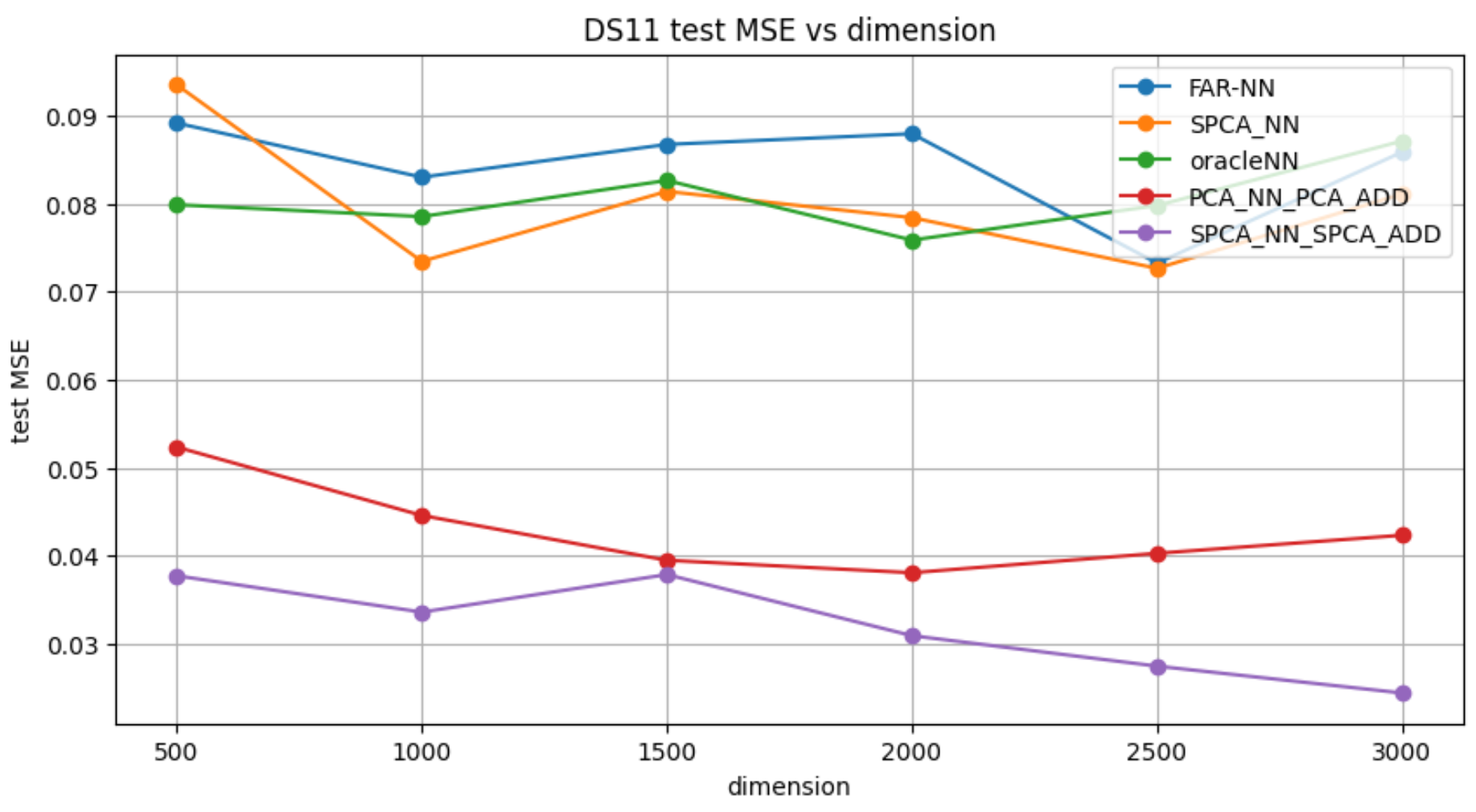}
    \caption*{(b)}
    \label{fig:ds02_table2}
\end{subfigure}
\caption{Dataset 1-1: test MSE against dimension. Figure (b) provides a zoomed-in view of the top-performing models.}
\label{fig:ds02_table}
\end{figure}



\begin{figure}[H]
\centering

\begin{subfigure} 
    \centering
    \includegraphics[width=0.95\textwidth]{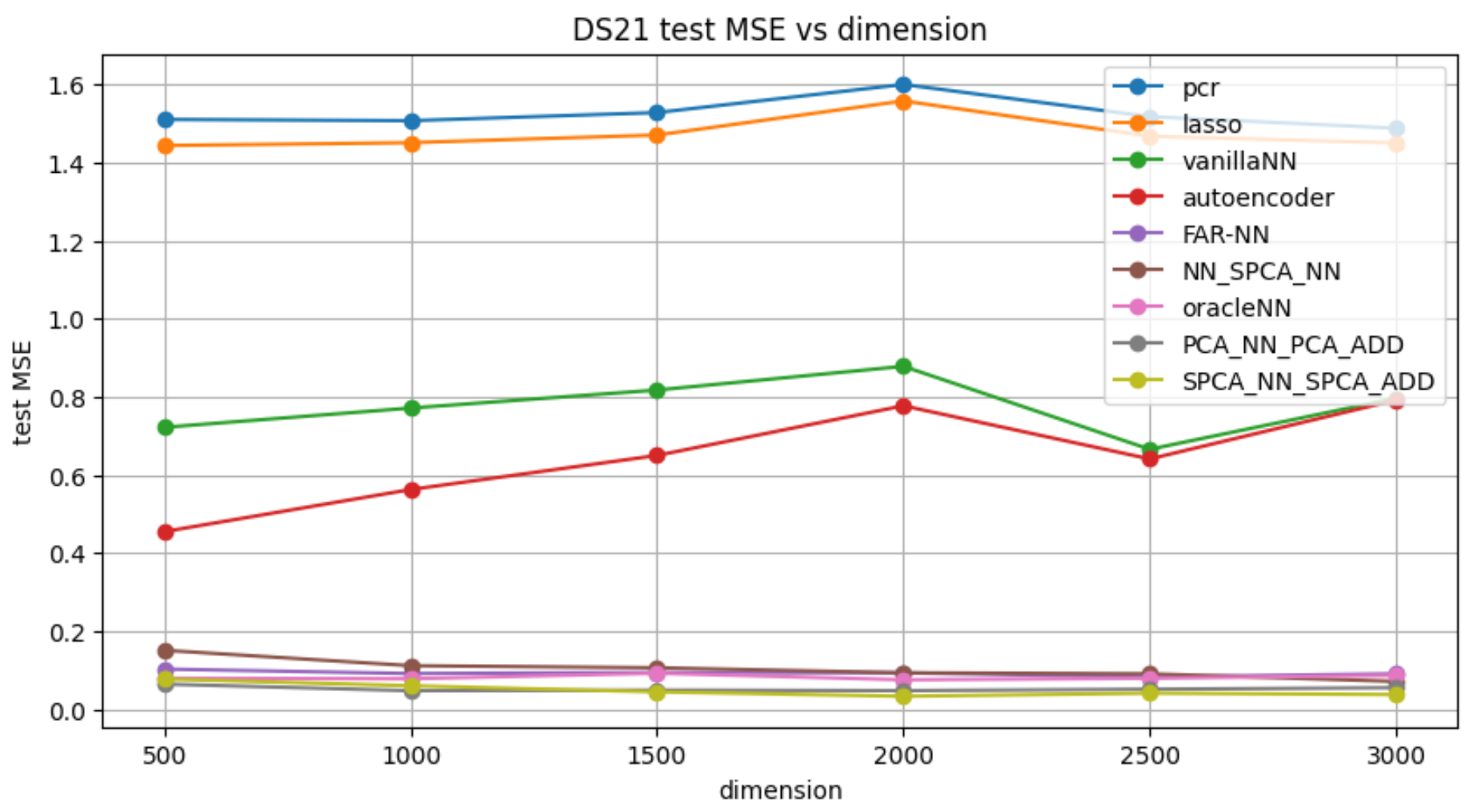}
    \caption*{(a)}
    \label{fig:ds12_table1}
\end{subfigure}
\hfill 

\begin{subfigure}
    \centering
    \includegraphics[width=0.95\textwidth]{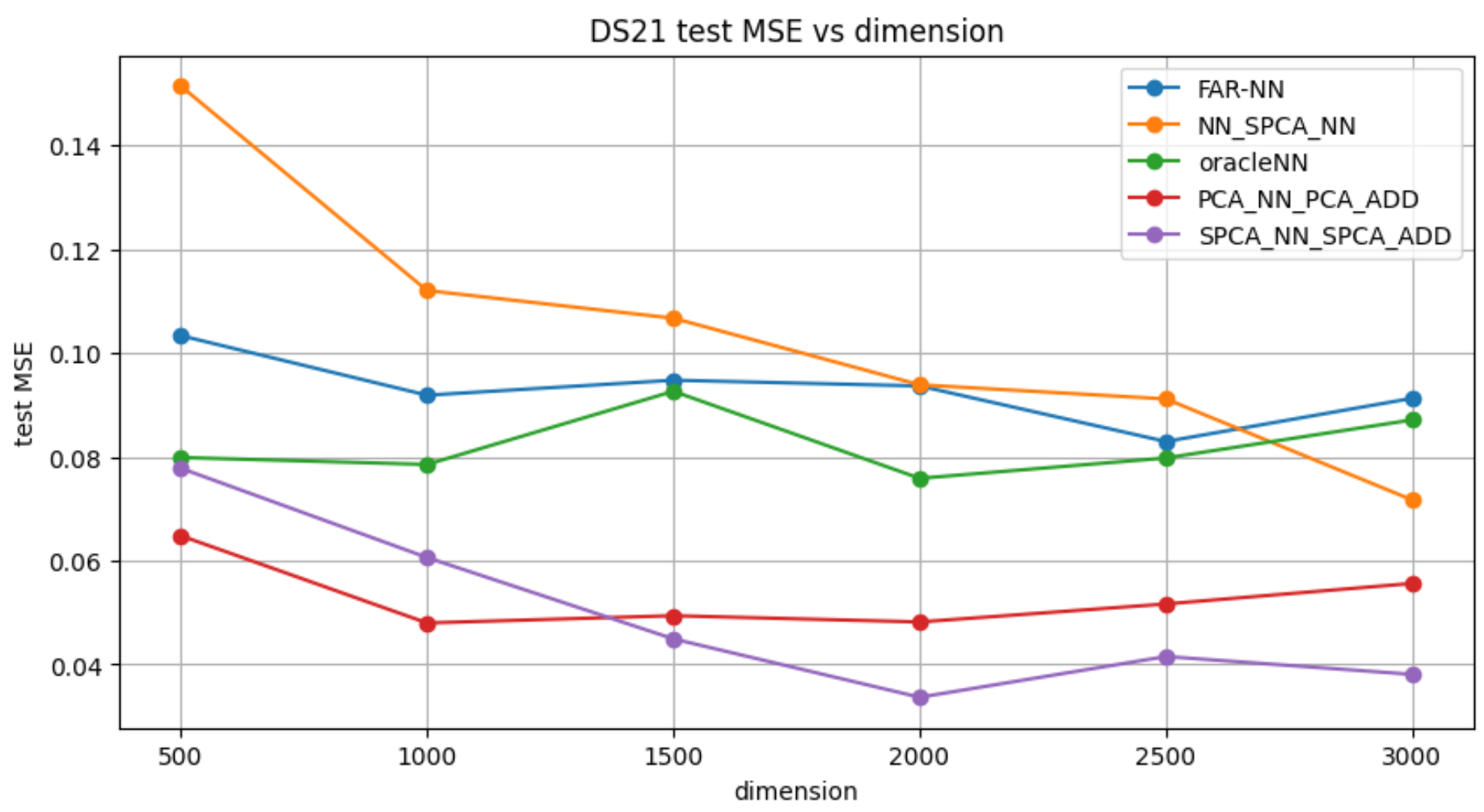}
    \caption*{(b)}
    \label{fig:ds12_table2}
\end{subfigure}
\caption{Dataset 2-1: test MSE against dimension. Figure (b) provides a zoomed-in view of the top-performing models.}
\label{fig:ds12_table}
\end{figure}



We vary the input dimension from \(p=500\) to \(p=3000\) and plot the test MSE against the dimension for all models. Figures \ref{fig:ds02_table} (a) and \ref{fig:ds12_table} (a) show that all neural network models outperform the classical PCR and Lasso models, while neural network models with a low-rank structure consistently outperform vanilla neural networks.

For Dataset 1-1, where the factor-observation relationship is linear, Figure \ref{fig:ds02_table} (b) demonstrates that our \(SPCA\_NN\) model performs similarly to or better than the FAR-NN and OracleNN models. In the case of Dataset 2-1, which exhibits a non-linear factor-observation relationship, we incorporate a soft PCA layer within the network architecture. As shown in Figure \ref{fig:ds12_table} (b), this approach achieves results comparable to FAR-NN and OracleNN when the number of dimensions \(p\) exceeds 1500.

Furthermore, for both data sets, the GFADNN model, structured as \( PCA\allowbreak\_NN\allowbreak\_PCA\allowbreak\_ADD \) and \( SPCA\allowbreak\_NN\allowbreak\_SPCA\allowbreak\_ADD \), where PCA and soft PCA operations are applied iteratively and fed into additive subnetworks, significantly reduces the test MSE across all dimensions. We also observe that models with soft PCA layer(s) exhibit a slight downward trend in test MSE as the dimensionality increases, while this trend is less clear for the vanillaNN and autoencoder models. This behavior is also documented in \citet{fan2023factor}, indicating that imposing a low-rank structure in neural network models, with the number of latent factors fixed, can even benefit from a reasonable and finite increase in dimension.

\section{Empirical Studies}
\label{sec:section5}
In this section, we evaluate our model by comparing it against benchmark neural network models and classical statistical models using real-world data sets. For each dataset, we construct a high-dimensional setting to perform regression tasks, with the primary focus on model performance comparison.

\subsection{Forecasting SPY ETF Price Returns}
An Exchange-Traded Fund (ETF) targeting specific sectors offers a focused investment approach, concentrating on companies within a particular industrial sector. In this experiment, we explore Sector ETFs from S\&P Global, specifically their S\&P 500 Sector ETFs. These ETFs break down the S\&P 500 into eleven distinct sector index funds. Our experiment aims to predict the next-day returns of S\&P 500 ETFs using data from all eleven Sector ETFs. The data, spanning from August 1, 2014, to August 1, 2024, was sourced from S\&P Global.

To forecast the next day's S\&P 500 return, we utilize a feature set comprising moving average returns over multiple time frames: 1, 5, 10, 20, and 126 days for each ETF. This approach results in a total feature dimensionality of 55. Our methodology employs a rolling training model, where we allocate 504 trading days—corresponding to the average number of trading days in two years—as our training window. For validation, we set aside 60 trading days, approximately equivalent to three months, followed by an additional 252 trading days designated for the testing phase.

The process begins with hyperparameter tuning to identify and select the optimal model configuration, which is based on achieving the lowest mean squared error (MSE) during the validation phase. The hyperparameter space is consistent with that used in the simulation experiment, except for the number of latent factors, which is allowed to vary between 11 and 22. Once selected, this model is trained on the training dataset, using the validation MSE as a criterion for early stopping to mitigate overfitting. The MSE obtained during the testing phase is then used to gauge the model's predictive accuracy.

To further evaluate our model's effectiveness, we implement a simple trading strategy based on the model’s forecasts. The forecasted signal is scaled by dividing it by 1.2 times the maximum value observed during the training period, then capped within the range [-1, 1]. This scaled signal is then used as the target percentage position. The profitability of this strategy is assessed by multiplying the position vector with the subsequent day’s S\&P 500 returns. 

\subsubsection{Portfolio Evaluation Metrics}
To evaluate our trading strategy, we use several performance metrics, including Directional Accuracy, Information Coefficient (IC), Sharpe Ratio, Turnover,  Maximum Drawdown and Average Percentage Position. We define each metric as follows:

1. \underline{Daily Returns}: The daily return \( r_t \) of the portfolio on day \( t \) is calculated as
   \[
   r_t = \frac{P_t - P_{t-1}}{P_{t-1}},
   \]
   where \( P_t \) is the portfolio value at the end of day \( t \).

2. \underline{Directional Accuracy} Directional Accuracy measures the percentage of days when the sign of the portfolio return aligns with the actual market direction. It is defined as
   \[
   \text{Directional Accuracy (Dir)} = \frac{\sum_{t=1}^{T} \mathbb{I} \left( \operatorname{sign}(r_t) = \operatorname{sign}(m_t) \right)}{T} \times 100\%,
   \]
   where \( \mathbb{I}(\cdot) \) is the indicator function, \( r_t \) is the portfolio return on day \( t \), \( m_t \) is the market return on day \( t \), and \( T \) is the total number of trading days.

3. \underline{Information Coefficient (IC)}: The Information Coefficient measures the correlation between the predicted returns and the actual returns, indicating the accuracy of the strategy's predictions. It is calculated as
   \[
   \text{IC} = \frac{\sum_{t=1}^{T} (f_t - \bar{f})(r_t - \bar{r})}{\sqrt{\sum_{t=1}^{T} (f_t - \bar{f})^2 \sum_{t=1}^{T} (r_t - \bar{r})^2}},
   \]
   where \( f_t \) is the predicted return, \( r_t \) is the actual return, \( \bar{f} \) and \( \bar{r} \) are the mean predicted and actual returns over the period.

4. \underline{Sharpe Ratio}: The Sharpe Ratio is used to evaluate the risk-adjusted return of the portfolio. It is defined as
   \[
   \text{Sharpe Ratio} = \frac{\mathbb{E}[r_t - r_f]}{\sigma_r},
   \]
   where \( r_t \) is the daily return, \( r_f \) is the risk-free rate, and \( \sigma_r \) is the standard deviation of daily returns.

5. \underline{Turnover}: Turnover measures the trading activity within the portfolio. It is calculated as
   \[
   \text{Turnover} = \frac{1}{T} \sum_{t=1}^{T} \sum_{i=1}^{N} \left| w_{i, t} - w_{i, t-1} \right|,
   \]
   where \( w_{i, t} \) is the portfolio weight of asset \( i \) at time \( t \), and \( N \) is the number of assets in the portfolio, and $N=1$ for this experiment.

6. \underline{Maximum Drawdown}: Maximum Drawdown represents the maximum observed loss from a peak to a trough in the portfolio's value, expressed as a percentage. It is defined as
   \[
   \text{Maximum Drawdown} = \max_{t \in [0, T]} \left( \frac{\max_{0 \le s \le t} P_s - P_t}{\max_{0 \le s \le t} P_s} \right) \times 100\%,
   \]
   where \( P_t \) is the portfolio value at time \( t \), and \( T \) is the end of the evaluation period.

7. \underline{Average Percentage Position}: Average Percentage Position represents the mean absolute position within the portfolio, calculated as the average of the absolute portfolio weights over the evaluation period. It is defined as

\[
\text{Average Percentage Position} = \frac{1}{T} \sum_{t=1}^{T} \frac{1}{N} \sum_{i=1}^{N} |w_{i,t}| \times 100\%,
\]
\noindent where \( w_{i,t} \) is the portfolio weight of asset \( i \) at time \( t \), \( N \) is the number of assets in the portfolio, and \( T \) is the end of the evaluation period.

Each of these metrics provides a different perspective on the portfolio's performance, helping us to assess returns, risk, accuracy of predictions, and trading costs comprehensively.

\subsubsection{Portfolio Performance Summary}
The outcomes are summarized in the table provided below.

\begin{table}[h!]
\centering
\resizebox{\textwidth}{!}{
\begin{tabular}{|l|r|r|r|r|r|r|r|}
\hline
\textbf{Model} & \textbf{Ret} & \textbf{Sharpe} & \textbf{MaxDD} & \textbf{Turnover} & \textbf{Dir} & \textbf{IC} & \textbf{AvgPos} \\ \hline
pcr & 0.061 & 0.528 & -0.150 & 0.232 & 0.509 & 0.045 & 0.327 \\
\hline
lasso & 0.082 & 0.436 & -0.234 & 0.057 & 0.534 & 0.017 & 0.919 \\
\hline
vanillaNN & 0.041 & 0.284 & -0.289 & 0.226 & 0.534 & 0.014 & 0.348 \\
\hline
autoencoder & 0.034 & 0.263 & -0.204 & 0.312 & 0.514 & 0.023 & 0.374 \\
\hline
FAR-NN & 0.073 & 0.523 & -0.185 & 0.320 & 0.521 & \textbf{0.059} & 0.445 \\
\hline
\textbf{PCA\_NN\_PCA\_ADD} & 0.080 & \textbf{0.767} & -0.159 & 0.150 & \textbf{0.539} & 0.057 & 0.286 \\
\hline
\textbf{SPCA\_NN\_SPCA\_ADD} & 0.139 & \textbf{1.244} & -0.099 & 0.133 & \textbf{0.543} & \textbf{0.071} & 0.495 \\
\hline
\end{tabular}
}
\caption{SP500 ETF Trading Strategies Performance Table, where Return and Sharpe ratio are annualized. }
\label{table:performance_metrics1}
\end{table}

The tables above show that our models achieve superior Sharpe ratios of 0.767 and 1.244, lower maximum drawdowns of -15.9\% and -9.9\%, and relatively lower turnover.

It is worth noting that the average percentage position of these portfolios is only around 30\%. To enhance the strategy, we decided to increase the position size and reduce turnover. First, we applied a 20-day rolling mean to smooth the position, followed by volatility targeting to achieve an annualized volatility of 20\%, similar to that of the S\&P 500 index. Finally, we capped the position within the range [-2, 2] to manage leverage. The resulting performance is presented in Table \ref{table:performance_metrics2}.

\begin{table}[h!]
\centering
\resizebox{\textwidth}{!}{
\begin{tabular}{|l|r|r|r|r|r|r|r|}
\hline
\textbf{Model} & \textbf{Ret} & \textbf{Sharpe} & \textbf{MaxDD} & \textbf{Turnover} & \textbf{Dir} & \textbf{IC} & \textbf{AvgPos} \\ \hline
pcr & 0.097 & 0.505 & -0.311 & 0.150 & 0.507 & 0.024 & 1.248 \\
\hline
lasso & 0.155 & 0.805 & -0.168 & 0.052 & 0.525 & 0.027 & 1.407 \\
\hline
vanillaNN & 0.124 & 0.652 & -0.330 & 0.116 & 0.531 & \textbf{0.041} & 1.238 \\
\hline
autoencoder & 0.062 & 0.318 & -0.414 & 0.151 & 0.510 & -0.012 & 1.266 \\
\hline
FAR-NN & 0.152 & 0.779 & -0.229 & 0.114 & 0.527 & 0.034 & 1.276 \\
\hline
\textbf{PCA\_NN\_PCA\_ADD} & 0.247 & \textbf{1.276} & -0.170 & 0.102 & 0.545 & \textbf{0.067} & 1.285 \\
\hline
\textbf{SPCA\_NN\_SPCA\_ADD} & 0.196 & \textbf{1.005} & -0.176 & 0.050 & \textbf{0.546} & 0.031 & 1.401 \\
\hline
\end{tabular}
}
\caption{Performance metrics for various models}
\label{table:performance_metrics2}
\end{table}

The tables above demonstrate that our model maintains a Sharpe ratio of 1.276 and 1.005 while enhancing the annualized return and significantly reducing turnover.

\subsection{FRED-MD Dataset}
In this section, we evaluate the performance of our FANN estimator against other high-dimensional linear estimators using the macroeconomics dataset FRED-MD \citep{mccracken2016fred}. FRED-MD has been used as one of the benchmarks for applying big data techniques, such as random subspace methods \citep{boot2019forecasting}, sufficient dimension reduction \citep{barbarino2017unified}, and various lasso-type regressions \citep{smeekes2018macroeconomic}. Smeekes and Wijler demonstrated that lasso-type estimators perform well with this dataset and are more robust to model misspecification than factor models, even when the underlying data-generating process has a factor structure. Additionally, research by \citet{medeiros2021forecasting} noted that the DeepNN model generally exhibits the poorest performance with this dataset. Consequently, this dataset poses significant challenges for our neural network model. 

The FRED-MD dataset comprises $p = 134$ monthly U.S. macroeconomic variables from January 1959, including metrics such as the unemployment rate and Treasuray bill rate. \citet{mccracken2016fred} demonstrate that these variables are largely describable through several latent factors. For each target response variable $y$, we utilize the vector of all other variables $\mathbf{x}_t$ to predict $y_t$ at each time index $t \in [T]$. This analysis demonstrates the ability of our estimator to capture relationships in high-dimensional datasets and generalize well in different scenarios.

The data spans from January 1980 to July 2022, processed according to \citet{mccracken2016fred}, resulting in a sample size of $n = 330$. Specifically, data from January 1980 to September 2009 (200 months, 60\% of the data) are used for training and validation, while data from October 2009 to July 2022 (130 months, denoted as $D_{\text{test}}$, 40\% of the data) are utilized for testing. Within the former period, a random sample of 70\% ($D_{\text{train}}$) is used for model training, and the remaining 30\% ($D_{\text{valid}}$) for validation and model selection. The performance of the estimator $\hat{m}_b$ is assessed using the out-of-sample $R^2$, defined as
\[
R^2_{\text{OOS}} = 1 - \frac{\sum_{(x,y) \in D_{\text{test}}} (\hat{m}(x) - y)^2}{\sum_{(x,y) \in D_{\text{test}}} (\bar{y}_{\text{train}} - y)^2} \quad \text{with} \quad \bar{y}_{\text{train}} = \sum_{(x,y) \in D_{\text{train}}} y.
\]

We begin with the model specification from \citet{fan2023factor}, setting the hyperparameter \( r \in \{4, 5\} \), depth \( L \in \{2, 3\} \), and width \( N = 32 \). This configuration is chosen to balance the moderate sample size \( n \) and ambient dimension \( p \). All other regularization hyperparameters ranges are kept consistent with those used in the simulation experiments.

Each variable is treated as a target variable in the regression, resulting in a total of 127 regression tasks. Performance was evaluated using the negative R-squared metric, with results reported for some selected variables for demonstration purposes, as shown in Table~\ref{table:FRED_model_performance}. Our analysis reveals that the GFANN model outperforms the Lasso model in 77 out of 127 cases, demonstrating a significant advantage. In comparison, FAST-NN surpasses the Lasso model in only 30 cases. These findings highlight the robustness and efficiency of the GFANN model in handling complex high-dimensional regression tasks relative to traditional methods.

To further compare the relative performance across models, we also record the relative rank of the test scores for each regression task, assigning a rank of 1 to the lowest (best) and 5 to the highest score. The average rank across all 127 regression tasks is summarized in Table~\ref{table:model_metrics}, showing that our GFANN model achieves the lowest average rank, even on a dataset that poses substantial challenges for neural network models.

\begin{table}[h!]
\centering
\begin{tabular}{|l|r|r|r|r|r|}
\hline
\textbf{target} & \textbf{lasso} & \textbf{pcr} & \textbf{vanillaNN} & \textbf{FAST-NN} & \textbf{GFANN} \\ \hline
CP3Mx         & -0.581 & 0.623 & 0.194 & 1.170 & -0.629 \\ \hline
USWTRADE      & -0.026 & -0.028 & 0.189 & 0.690 & -0.136 \\ \hline
FEDFUNDS      & -0.377 & 0.396 & 1.847 & 0.250 & -0.471 \\ \hline
CES2000000008 & -0.782 & 0.030 & -0.006 & -0.162 & -0.776 \\ \hline
CUSR0000SAS   & -0.812 & 0.265 & 0.475 & -0.273 & -0.875 \\ \hline
EXUSUKx       & -0.474 & 0.053 & -0.030 & 0.149 & -0.431 \\ \hline
UEMP5TO14     & -0.428 & 0.075 & 0.045 & 0.078 & -0.415 \\ \hline
TOTRESNS      & -0.767 & 0.013 & -0.164 & -0.300 & -0.773 \\ \hline
ACOGNO        & -0.418 & -0.241 & -0.227 & 0.043 & -0.426 \\ \hline
EXJPUSx       & -0.321 & 0.190 & 0.073 & 0.277 & -0.156 \\ \hline
\end{tabular}
\caption{Comparison of model performance for selected targets; the metric employed here is the negative $R^2_{\text{OOS}}$.}
\label{table:FRED_model_performance}
\end{table}

\begin{table}[h!]
\centering
\begin{tabular}{|l|r|r|r|r|r|}
\hline
\textbf{lasso} & \textbf{pcr} & \textbf{vanillaNN} & \textbf{FAST-NN} & \textbf{GFANN} \\ \hline
1.961 & 4.378 & 4.0 & 2.89 & 1.772 \\ \hline
\end{tabular}
\caption{Average rank across all 127 regression tasks, with a rank of 1 assigned to the lowest (best) test score and 5 to the highest. Lower ranks indicate better performance.}
\label{table:model_metrics}
\end{table}

\section{Conclusion}
\label{sec:section6}
This study has addressed the challenges involved in analyzing high-dimensional data sets, particularly those embodying latent low-dimensional structures concealed by non-linear, and noise-laden relationships. Our novel 
strategy enhances neural networks through the incorporation of additive and factor layers, which can be effortlessly embedded at any step  within the neural network architecture. This flexibility proves especially effective in managing hierarchical and compositional data. A notable finding is that the sequential arrangement of additive layers following factor layers constitutes a highly efficient architecture for factor additive data-generating processes, showcasing superior performance with minimal parameter requirements and robustness against variations in architecture-related hyperparameters. Beyond integrating additive and factor models within neural networks seamlessly, our research validates the effectiveness of these innovations through comprehensive simulation studies and practical applications to the prediction of SPY ETF price returns and FRED-MD macroeconomic indicators.


\newpage
During the preparation of this work, the author(s) used Chatgpt in order to polish the writing. After using this tool/service, the author(s) reviewed and edited the content as needed and take(s) full responsibility for the content of the publication.

\bibliographystyle{elsarticle-num-names} 

\newpage
\section{Appendices}

\subsection{Table of Notations and Acronyms} \label{sec:table_notation}
\begin{table}[h!]
\centering
\begin{tabular}{|l|p{0.65\textwidth}|}
\hline
\textbf{Notation/Acronym} & \textbf{Description} \\ \hline
$k$ & number of latent factors \\ \hline
$p$ & input dimension \\ \hline
PCA & Principal Component Analysis \\ \hline
PCR & Principal Component Regression \\ \hline
vanillaNN & A vanilla ReLU-Neural Network model \\ \hline
oracleNN & A vanilla ReLU-Neural Network model fed with \textbf{latent factors} \\ \hline
FAR-NN & Factor Augmented Regression Using Neural Network \\ \hline
FAST-NN & Factor Augmented Sparse Throughput Neural Network \\ \hline
GFNN & Generalized Factor Neural Network Model \\ \hline
GFADNN & Generalized Factor Additive Neural Network Model, GFNN with additive layers. \\ \hline
GFANN & Generalized Factor Augmented Neural Network Model, GFNN with augmented residual components. \\ \hline
$PCA\_NN$ & A GFNN model with the PCA layer as the first layer, followed by standard layers. \\ \hline
$SPCA\_NN$ & A GFNN model with the Soft PCA layer as the first layer, followed by standard layers. \\ \hline
$PCA\_NN\_PCA\_ADD$ & A GFADNN model with a PCA layer, standard layer(s), another PCA layer, then followed by additive layer(s). \\ \hline
$PCA\_NN\_ADD\_PCA$ & A GFADNN model with a PCA layer, standard layer(s), additive layer(s), then followed by another PCA layer. \\ \hline
\end{tabular}
\caption{Table of Notations and Acronyms}
\label{table:notation_acronyms}
\end{table}

\newpage

\subsection{Rationale for Model Comparison in Simulation Studies} \label{sec:model_comparison}
There is a substantial body of literature that compares neural network models using identical hyperparameter settings. However, hyperparameter settings are highly specific to each model, and therefore, for a fair comparison, we aim to select the best hyperparameter setting for each individual model. To achieve this, we systematically vary crucial hyperparameters, such as learning rates, network width, network depth, and batch size. Upon establishing the search space, hyperparameter optimization is conducted using Optuna, with the adoption of the Tree-structured Parzen Estimator (TPE) algorithm—a Bayesian optimization technique—for the tuning process. The configuration that results in the minimal validation error is selected for each model. Subsequently, we vary the random seeds to generate over 20 random samples. For each sample, models are trained, the validation data is used for early stopping, and test errors are recorded. Ultimately, model performance is compared based on their average test error. In particular, the performance of the estimator $\hat{m}(x)$ is evaluated via the empirical mean squared error.

\subsection{Hyperparameter Sensitivity}

\begin{figure}[H]
\centering

\begin{subfigure} 
    \centering
    \includegraphics[width=0.95\textwidth]{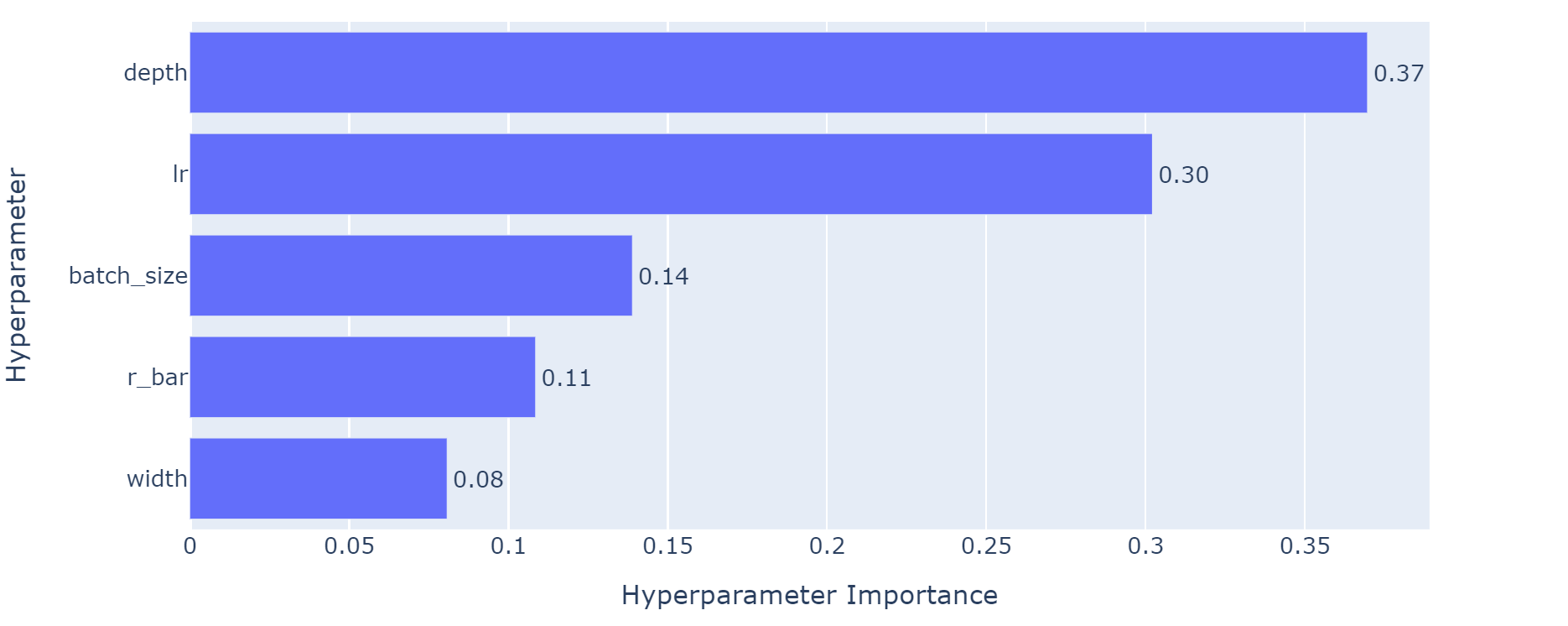}
    \caption*{(a) FAR-NN}
    \label{subfig:hyper_importance_FAR-NN}
\end{subfigure}
\hfill 

\begin{subfigure}
    \centering
    \includegraphics[width=0.95\textwidth]{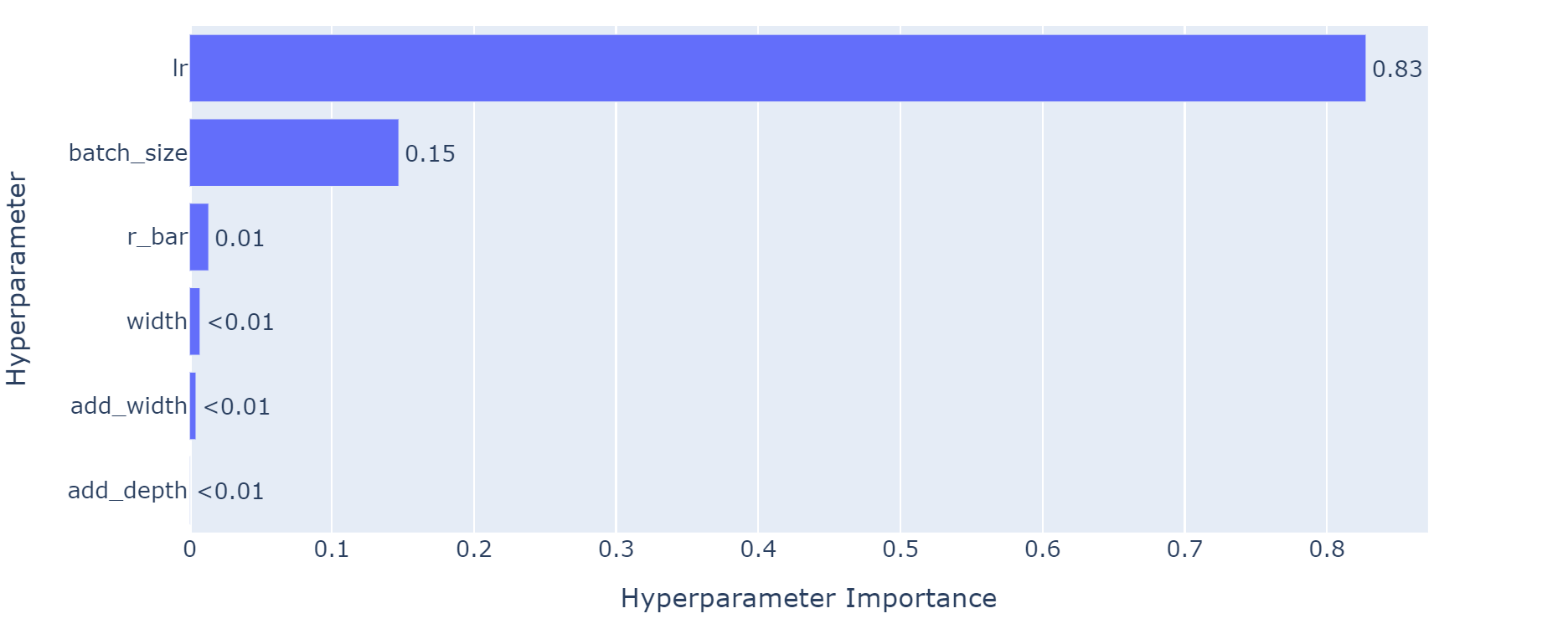}
    \caption*{(b) PCA\_NN\_PCA\_ADD}
    \label{subfig:hyper_importance_PCA_NN_PCA_ADD}
\end{subfigure}

\vspace{0.5cm} 

\begin{subfigure} 
    \centering
    \includegraphics[width=0.95\textwidth]{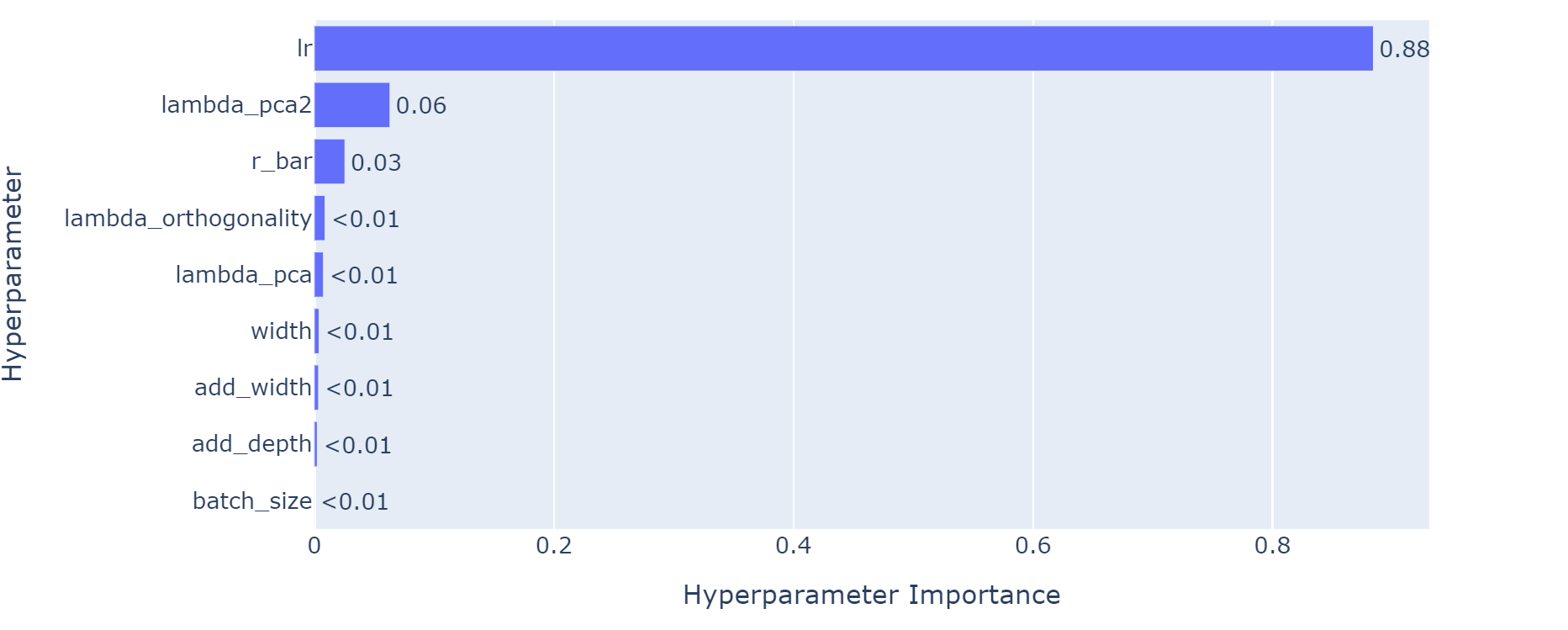}
    \caption*{(c) SPCA\_NN\_SPCA\_ADD}
    \label{subfig:hyper_importance_SPCA_NN_SPCA_ADD}
\end{subfigure}

\caption{Hyperparameter Importances diagram of (a) FAR-NN, (b) PCA\_NN\_PCA\_ADD, and (c) SPCA\_NN\_SPCA\_ADD. This presents the relative importance of the hyperparameters tuned in the training process}
\label{fig:hyper_importance}
\end{figure}

Through the hyperparameter tuning process, we gain insights into the significance of various hyperparameters for each model type, specifically focusing on \(FAR\text{-}NN\), \(PCA\_NN\_PCA\_ADD\), and \(SPCA\_NN\_SPCA\_ADD\). By evaluating the relative importance of these hyperparameters, with the learning rate as a reference point, we observe that the learning rate has larger relative importance in Figure~\ref{fig:hyper_importance}(b) \(PCA\_NN\_PCA\_ADD\) and Figure~\ref{fig:hyper_importance}(c) \(SPCA\_NN\_SPCA\_ADD\). In contrast, in Figure~\ref{fig:hyper_importance}(a) \(FAR\text{-}NN\), hyperparameters such as batch size and architecture-related features, including width and depth, exhibit comparable influence to the learning rate. This suggests a heightened sensitivity of the model to architectural adjustments, relative to learning rate changes.

Conversely, for \(PCA\_NN\_PCA\_ADD\) and \(SPCA\_NN\_SPCA\_ADD\), architecture-related hyperparameters demonstrate less significance relative to the learning rate, providing insights into the robustness of this model's architecture.

\subsection{Ablation study}

\begin{figure}[H]
    \centering
    \includegraphics[width=\columnwidth]{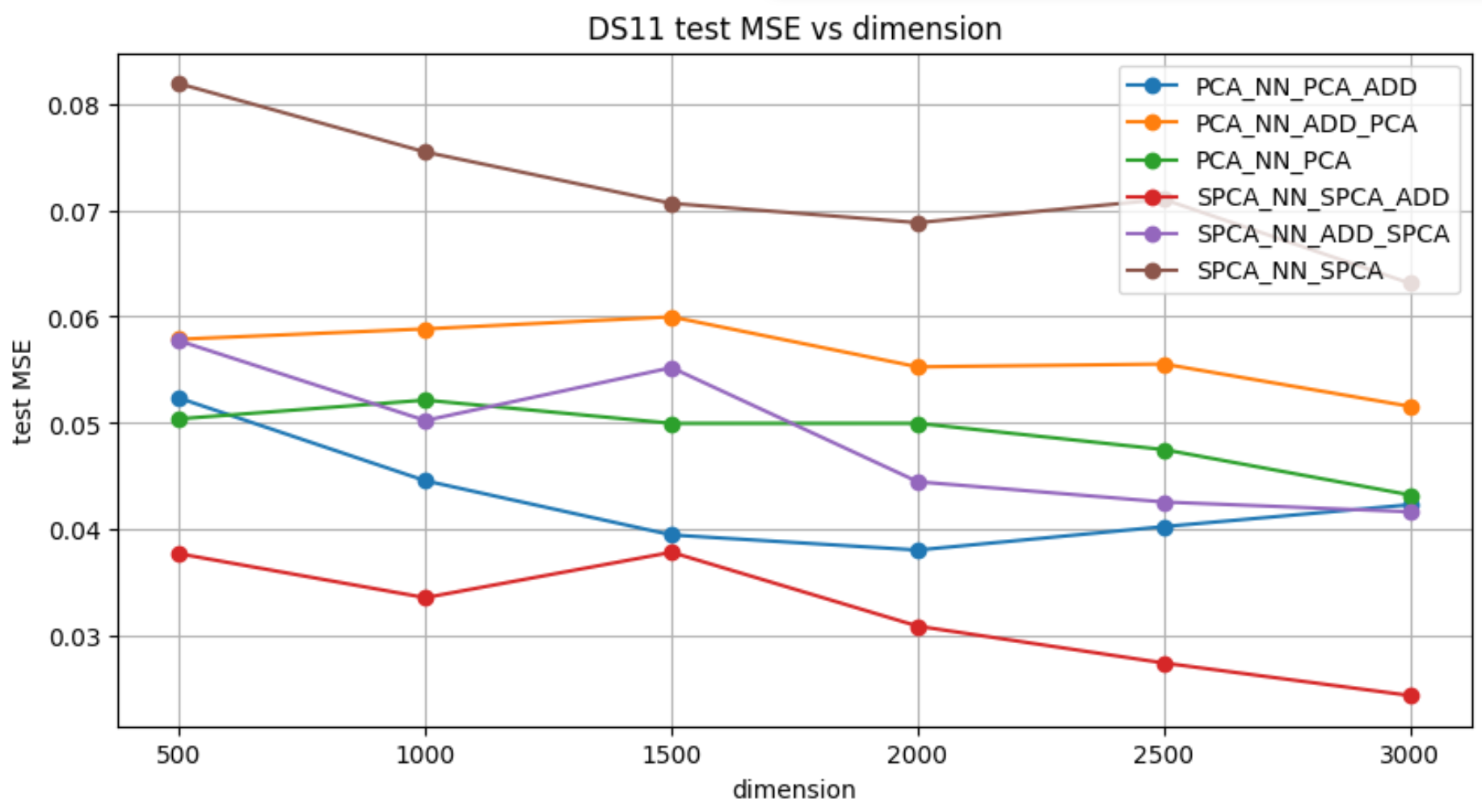}
    \caption{test mse against dimension when observations follow a linear factor model. The benchmark models are PCA\_NN\_PCA\_ADD in blue and SPCA\_NN\_SPCA\_ADD in red.}
    \label{fig:ablation1}
\end{figure}


\begin{figure}[H]
    \centering
    \includegraphics[width=\columnwidth]{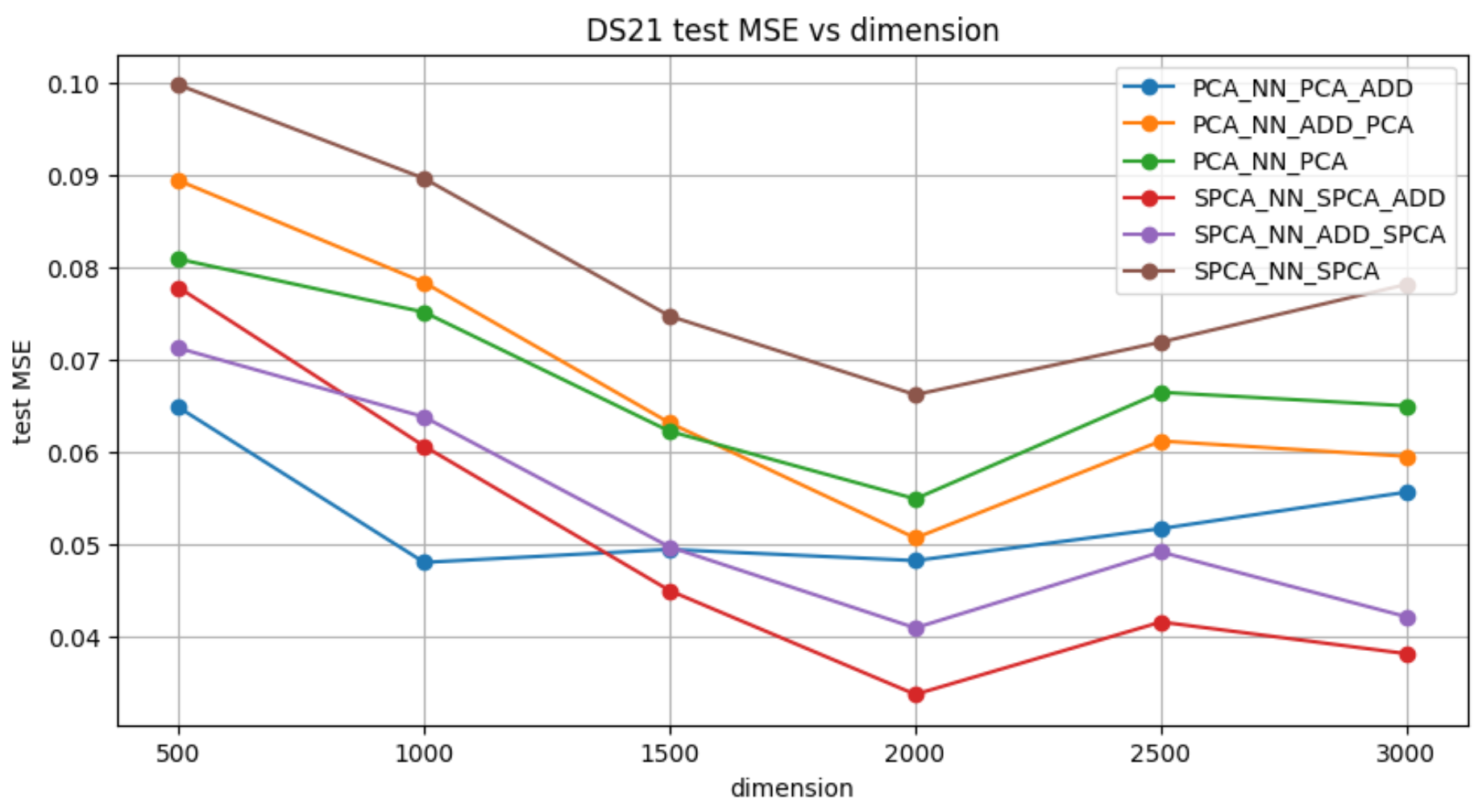}
    \caption{test mse against dimension when observations follow a non-linear factor model. The benchmark models are PCA\_NN\_PCA\_ADD in blue and SPCA\_NN\_SPCA\_ADD in red.}
    \label{fig:ablation2}
\end{figure}

In this section, we evaluate the performance implications of modifying key components within our GFADNN model. Specifically, we examine the impact of replacing the Additive layer with a simple linear layer, resulting in the models \( PCA\_NN\_PCA \) and \( SPCA\_NN\_SPCA \). Additionally, we investigate the effect of swapping the second (Soft) PCA layer with the Additive layers, producing the models \( PCA\_NN\_ADD\_PCA \) and \( SPCA\_NN\_ADD\_SPCA \). Figures \ref{fig:ablation1} and \ref{fig:ablation2} display the average test MSE of these models across 20 random seeds for different dimensions. The results indicate that either removing the Additive layers or changing their order negatively impacts performance, highlighting their significance in the model's architecture.


\subsection{An illustration of additive layers} \label{sec:additive_layer_graph}
\begin{figure}[H]
    \centering
    \includegraphics[width=\columnwidth]{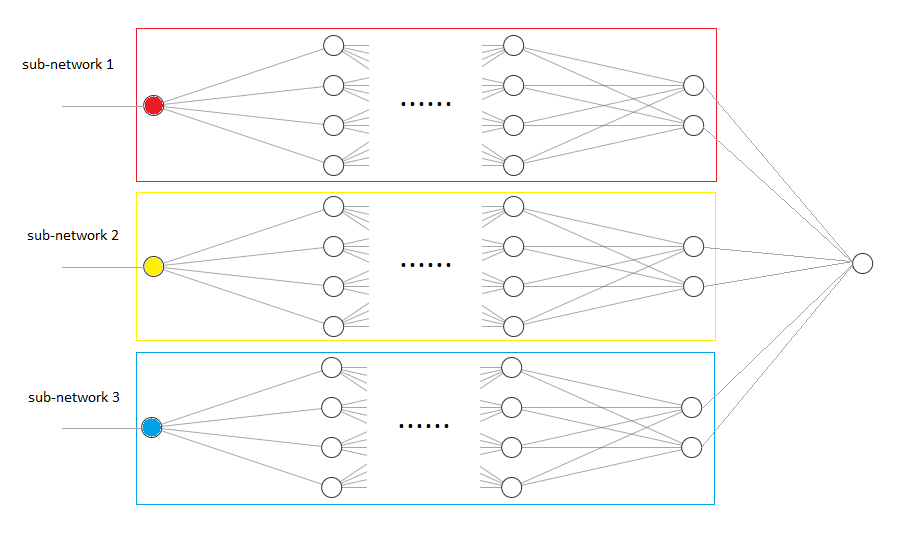}
    \caption{Additive networks, without interactions among neurons across different sub-networks.}
\label{fig:additive_layer}
\end{figure}
\subsection{Computational graph involving the PCA layer}
\label{sec:computation_graph}
\begin{figure}[!h]
\centering
    \includegraphics[height = 21cm, clip ]{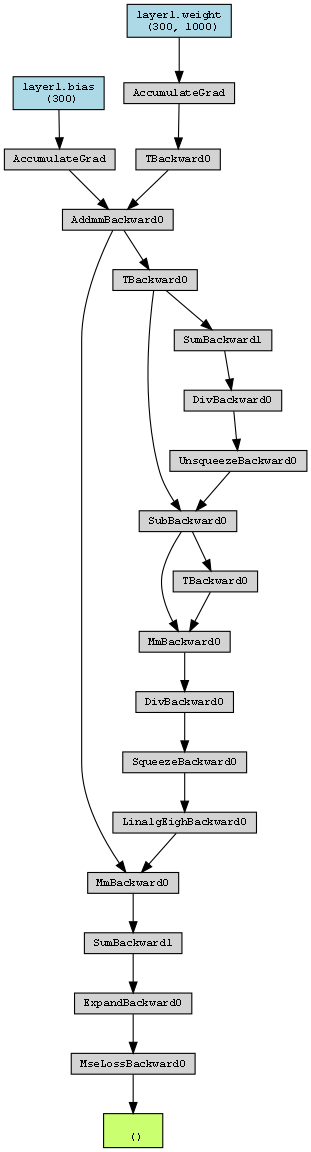}
    \caption{An illustration of the computational graph, highlighting the tracking of gradients during a PCA operation.}
\label{fig:loss_nontorch}
\end{figure}

\end{document}